%% file: neurips_2025.tex
\documentclass{article}

\usepackage[preprint]{neurips_2025}

\usepackage{amsmath}
\usepackage{amssymb}

\usepackage{tabularx,booktabs,xcolor}
\usepackage{adjustbox}   % for adjustbox
\usepackage{booktabs}    % for \toprule etc.
\usepackage{array}
\newcolumntype{C}[1]{>{\centering\arraybackslash}p{#1}}
\newcolumntype{C}[1]{>{\centering\arraybackslash}p{#1}}
\newcolumntype{Y}{>{\centering\arraybackslash}X}  % centered X columns
\usepackage[utf8]{inputenc} % allow utf-8 input
\usepackage[T1]{fontenc}    % use 8-bit T1 fonts
\usepackage{hyperref}       % hyperlinks
\usepackage{url}            % simple URL typesetting
\usepackage{amsfonts}       % blackboard math symbols
\usepackage{nicefrac}       % compact symbols for 1/2, etc.

\usepackage{colortbl}
\usepackage{xcolor}
\definecolor{gold}{RGB}{255,215,0}
\definecolor{silver}{RGB}{192,192,192}
\definecolor{bronze}{RGB}{205,127,50}
\usepackage{microtype}      % microtypography       % colors
\usepackage{lipsum}
\usepackage{enumitem}
\usepackage{natbib}
\usepackage{array}

\usepackage{amsmath}
\usepackage[table,dvipsnames]{xcolor}  % dvipsnames 可选，用于更多命名颜色
\usepackage{booktabs}                  % for \toprule, \midrule, \bottomrule
\usepackage{array}                     % for \newcolumntype
\newcolumntype{C}[1]{>{\centering\arraybackslash}p{#1}}
\title{EvoEmpirBench: Dynamic Spatial Reasoning with Agent-ExpVer}

\author{%
Pukun Zhao\textsuperscript{1}\thanks{Equal contribution}\\
\textsuperscript{1}Guangdong University of Finance and Economics\\
\texttt{zhaopukun@student.gdufe.edu.cn}
\And
Longxiang Wang\textsuperscript{2}\footnotemark[1]\\
\textsuperscript{2}Chongqing University\\
\texttt{longxiangwang@stu.cqu.edu.cn}
\And
Miaowei Wang\textsuperscript{4}\footnotemark[1]\\
\textsuperscript{4}University of Edinburgh\\
\texttt{m.wang-123@sms.ed.ac.uk}
\AND
Chen Chen\textsuperscript{1}\\
\texttt{Allen821@student.gdufe.edu.cn} 
\And
Fanqing Zhou\textsuperscript{1}\\
\texttt{zhoyfanqing@gmail.com}
\And
Haojian Huang\textsuperscript{3}\thanks{Corresponding author}\\
\textsuperscript{3}The University of Hong Kong\\
\texttt{haojianhuang@connect.hku.hk}
}

\begin{document}

\maketitle

\begin{abstract}
 Most existing spatial reasoning benchmarks focus on static or globally observable environments, failing to capture the challenges of long-horizon reasoning and memory utilization under partial observability and dynamic changes. We introduce two dynamic spatial benchmarks—locally observable maze navigation and match-2 elimination—that systematically evaluate models' abilities in spatial understanding and adaptive planning when local perception, environment feedback, and global objectives are tightly coupled. Each action triggers structural changes in the environment, requiring continuous update of cognition and strategy. We further propose a subjective experience-based memory mechanism for cross-task experience transfer and validation. Experiments show that our benchmarks reveal key limitations of mainstream models in dynamic spatial reasoning and long-term memory, providing a comprehensive platform for future methodological advances. Our code and data are available at the \textcolor{magenta}{\href{https://anonymous.4open.science/r/EvoEmpirBench-143C/README.md}{GitHub Repository}}. Please also refer to the supplementary material for our video demonstration.

\end{abstract}

\input{files/1-introduction}
\input{files/2-related_work}

\input{files/3-benchmark}
\input{files/4-methodology}
\input{files/5-experiment}

\input{files/6-conclusion}

\bibliographystyle{plain}
\bibliography{neurips_2025}

%%%%%%%%%%%%%%%%%%%%%%%%%%%%%%%%%%%%%%%%%%%%%%%%%%%%%%%%%%%%

\input{files/appendix}

%%%%%%%%%%%%%%%%%%%%%%%%%%%%%%%%%%%%%%%%%%%%%%%%%%%%%%%%%%%%
% \input{files/checklist}

\end{document}

%% file: files/1-introduction.tex
\section{Introduction}
\label{sec:intro}
\begin{flushright}
``What we see changes what we know. What we know changes what we see.''\
\textemdash\ \emph{Jean Piaget}~\cite{piaget_construction_1954}
\end{flushright}

Over the past few years, Large Language Models (LLMs)\cite{chowdhery2023palm,zhang2022opt,zeng2022glm,touvron2023llama,achiam2023gpt,touvron2023llama2,chen2024bovila} have achieved impressive results across a wide range of static natural language benchmarks—including machine translation\cite{hendy2023good,xu2024contrastive,zhu2023multilingual,wang2023document,enis2024llm}, question answering~\cite{yao2023react,schick2023toolformer,shen2023hugginggpt,Madaan2023SelfRefineIR,Zhang2023DataCopilotBB,zhang2022multi,zhang2023expression,chen2024uncertainty,huang2025vistadpo,li2025text,zheng2024videogen}, and code generation~\cite{hong2023metagpt,wu2023autogen,du2025dependeval}—by leveraging massive pretraining corpora and ever-larger context windows.

However, evaluating genuine reasoning and adaptability in LLMs, especially in dynamic, interactive scenarios, remains a significant challenge. Existing benchmarks for reasoning~\cite{hao2024llm,srivastava2022beyond,valmeekam2023planbench,saparov2022language,hong2023metagpt,wu2023autogen,cobbe2021training} predominantly rely on static datasets, rendering them vulnerable to data contamination, where models overfit to training patterns and are prone to rapid performance saturation due to their limited scope~\cite{sainz2023nlp,white2024livebench,kiela2021dynabench}. While dynamic human-in-the-loop evaluations such as Chatbot Arena~\cite{chiang2024chatbot} bring fresh perspectives, they tend to capture subjective user preferences rather than objective reasoning skills~\cite{hu2024gamearena,li2024does}. Furthermore, Chatbot Arena struggles to evaluate specific reasoning capabilities, such as inductive and deductive reasoning, due to its reliance on unconstrained user inputs and the lack of precise control over the prompts' reasoning demands.

% Furthermore, these evaluations lack fine-grained control over specific reasoning tasks \MW{write example}, making it difficult to assess competencies like inductive and deductive reasoning in a targeted manner.

Recent work~\cite{chiang2024chatbot,zheng2023judging,zhao2024auto,wu2023smartplay,leng2025crosswordbench,hu2024gamearena,zhang2024agent,chen2024automanual} has begun exploring dynamic interactive tasks, but with limited generality. Many LLM-based agents excel only in narrowly defined domains and depend heavily on handcrafted prompts~\cite{zhang2024agent}. For example, the Agent-Pro framework~\cite{zhang2024agent} enables agent learning in games such as Blackjack and Texas Hold’em, but these settings feature relatively shallow interaction structures and limited information complexity, hindering transfer to more realistic and complex tasks~\cite{zhang2024agent}, such as navigation in unstructured environments~\cite{kober2013reinforcement} or real-time resource management~\cite{mao2016resource}.

To address these deficits, we introduce a suite of dynamic and interactive benchmark environments that more faithfully capture the complexity and uncertainty of real-world reasoning problems. Our benchmarks include a maze game and a match-2 game that require multi-step spatial reasoning, tool use, strategic planning, and adaptive behavior under partial observability. Unlike prior work, our environments feature dynamic elements—such as discoverable tools or obstacles, score-based resource management, and non-stationary information structures—demanding agents to continually gather information, revise plans, and generalize learned strategies.

Given these dynamic settings, traditional static “collect-then-repeat” training paradigms~\cite{mnih2015human,haarnoja2018soft,schulman2017proximal} are ill-suited. In contrast, human learning involves continual cycles of abstraction and rule induction, where experiences are retrieved, refined, and integrated into transferable knowledge as new situations arise~\cite{lake2017building,brown2020language,tenenbaum2011grow}. Inspired by this, we propose a cognitively grounded online learning framework with three collaborating agents: the \textit{GeoLink Agent} actively engages with the environment, the \textit{InsightForce Agent} abstracts experiences and distills transferable rules (“truths”), and the \textit{TruthWeaver Agent} manages these truths by merging, de-duplicating, and integrating new rules into structured memory. This architecture enables continual, adaptive learning without requiring offline data collection or post-deployment model updates.
\textbf{Our contributions are as follows:}
\begin{itemize}[leftmargin=*]
\item \textbf{Dynamic Reasoning Benchmark:} We present interactive, partially observable environments (maze and match-2 games) designed to rigorously test spatial, linguistic, and adaptive reasoning beyond static and narrow-domain evaluations (Section ~\ref{sec:Benchmark}).
\item \textbf{Cognitively-inspired Online Learning:} We propose a novel human-inspired learning framework with memory abstraction and rule distillation, enabling continual, parameter-free adaptation and lifelong learning in dynamic environments~ (Section ~\ref{sec:methods}).
\end{itemize}

%% file: files/2-related_work.tex
\section{Related Work}
\label{sec:related_work}
\textbf{Benchmarks for LLM Reasoning.}\quad The assessment of LLM reasoning has evolved from static to dynamic benchmarks. Early static benchmarks—covering logical reasoning~\cite{hao2024llm,srivastava2022beyond,valmeekam2023planbench,saparov2022language,hu2024recent}, coding~\cite{hong2023metagpt,wu2023autogen}, and mathematics~\cite{cobbe2021training}—probed LLMs using fixed datasets but struggled with data contamination~\cite{sainz2023nlp,white2024livebench} and performance saturation~\cite{kiela2021dynabench}. To address these limitations, dynamic evaluation approaches have emerged. Methods involving human or LLM judges~\cite{chiang2024chatbot,zheng2023judging,zhao2024auto} assess open-ended responses, but can introduce subjective biases~\cite{chen2024humans,li2024does,huang2024crest,huang2024evidential,ma2024beyond,huang2025trusted}. Game-based benchmarks provide a more robust and interactive evaluation of real-world problem-solving (see Table~\ref{tab:game_benchmarks}). SmartPlay~\cite{wu2023smartplay} and CrossWordBench~\cite{leng2025crosswordbench} leverage static environments to study spatial and multimodal reasoning, but lack dynamic, real-time interaction. GameArena~\cite{hu2024gamearena} incorporates human-LLM play in games like Akinator and Bluffing, yet does not provide fine-grained control over environmental dynamics or specific reasoning skills. Dynamic settings such as Agent-Pro~\cite{zhang2024agent}, focused on game theory in Blackjack and Poker, and AutoManual~\cite{chen2024automanual}, which measures task completion, are limited by the scope and diversity of scenarios. In contrast, our benchmark introduces fully dynamic, interactive game environments that challenge LLMs with multi-dimensional reasoning tasks—including spatial analysis, language understanding, and adaptive decision-making—under uncertainty. Continuous exploration and active agent-environment interaction better reflect real-world complexity, requiring long-horizon reasoning across local and global objectives.

\begin{table}[ht]
  \centering
\caption{%
  \textbf{Comparison of Game-Based Benchmarks for LLM Reasoning.} \textsuperscript{†}Agent perceives and is influenced by the environment, but actions do not modify it; \textsuperscript{‡}Agent perceives and modifies the environment, resulting in mutual influence; 
PO: Partially Observable; GLH: Global Long-Horizon.
}
  \label{tab:game_benchmarks}

  {\scriptsize                           % 全表用更小字号
  \rowcolors{2}{gray!10}{white}
  \begin{adjustbox}{max width=\textwidth,center}
    \begin{tabular}{%
        C{2.2cm}  % Benchmark
        C{1.3cm}  % # Types / Trials
        C{2.8cm}  % Reasoning Focus
        C{1.0cm}  % Dynamic
        C{1.8cm}  % Env Interaction
        C{1.8cm}  % Real-World Complexity
        C{1.3cm}  % Observability
      }
      \toprule
      \rowcolor{gray!30}
      \textbf{Benchmark}
        & \textbf{\# Types / Trials}
        & \textbf{Reasoning Focus}
        & \textbf{Dynamic}
        & \textbf{Env Interaction}
        & \textbf{Real-World Complexity}
        & \textbf{Observability} \\
      \midrule
      SmartPlay~\cite{wu2023smartplay}
        & 6 / 180
        & Spatial \& Planning
        & No
        & Yes†
        & Low
        & Full \\
      GameArena~\cite{hu2024gamearena}
        & 3 / 2000+
        & Deductive \& Inductive
        & No
        & No
        & Medium
        & Full \\
      CrossWordBench~\cite{leng2025crosswordbench}
        & 1 / 350
        & Multimodal
        & No
        & No
        & Low
        & Full \\
      Agent-Pro~\cite{zhang2024agent}
        & 2 / 187
        & Game-Theoretic
        & Yes
        & Yes‡
        & Medium
        & Full \\
      AutoManual~\cite{chen2024automanual}
        & 2 / 800+
        & Task Completion
        & Yes
        & Yes‡
        & Medium
        & Full \\
      \rowcolor{yellow!20}
      \textbf{Ours}
        & \textbf{2 / 180}
        & \textbf{Spatial \& GLH}
        & \textbf{Yes}
        & \textbf{Yes‡}
        & \textbf{High}
        & \textbf{PO} \\
      \bottomrule
    \end{tabular}
  \end{adjustbox}
  } % end scriptsize
\end{table}

% \begin{table}[h]
%     \centering
%     \caption{\textbf{Comparison of Game-Based Benchmarks for LLM Reasoning Evaluation} (*: Agent acts, Env responds statically; **: Agent acts, Env changes and responds, mutual influence; PO: Partially Observable)}
%     \label{tab:game_benchmarks}
%     \scriptsize
%     \begin{tabular}{>{\centering\arraybackslash}p{2.0cm} >{\centering\arraybackslash}p{1.7cm} >{\centering\arraybackslash}p{3.0cm} >{\centering\arraybackslash}p{0.6cm} >{\centering\arraybackslash}p{1.7cm} >{\centering\arraybackslash}p{2.5cm}}
%         \toprule
%         \textbf{Benchmark} & \textbf{Types/Sizes} & \textbf{Reasoning Focus} & \textbf{Dynamic} & \textbf{Env Interaction} & \textbf{Real-World Complexity} \\
%         \midrule
%         SmartPlay~\cite{wu2023smartplay} & 6/180 trials & Spatial, Planning & No & Yes* & Low \\
%         GameArena~\cite{hu2024gamearena} & 3/2000+ matches & Deductive, Inductive & No & No & Medium \\
%         CrossWordBench~\cite{leng2025crosswordbench} & 1/350 puzzles & Multimodal & No & No & Low \\
%         Agent-Pro~\cite{zhang2024agent} & 2/187 games & Game-Theoretic & Yes & Yes** & Medium \\
%         AutoManual~\cite{chen2024automanual} & 2/800+ tasks & Task Completion & Yes & Yes** & Medium \\
%         Ours & 2/180 levels & Spatial, Global Long-horizon & Yes & Yes** & High (PO) \\
%         \bottomrule
%     \end{tabular}
% \end{table}
% 2.关于智能体的持续学习机制：从回顾保持模型记忆以及在线学习等缓解灾难遗忘的方法论进行展开（最后的部分可以衔接automanual和agentPro）
\textbf{Continual Learning Mechanisms for Agents.}\quad
Continual learning (CL)\cite{wang2024CL} enables intelligent agents to incrementally adapt to new tasks while minimizing catastrophic forgetting\cite{kirkpatrick2017overcoming}. Recent work on CL for LLM-based agents~\cite{ma2021continual,bang2021rainbow,zhang2024agent,chen2024automanual,ghunaim2023real,kim2024online} emphasizes memory management and online learning, often drawing from cognitive principles. \textit{Memory management} is critical for retaining knowledge across tasks. Inspired by human episodic memory, experience replay methods store past interactions for periodic retraining~\cite{rolnick2019experience}. For example, CLIN~\cite{majumder2023clin} continually updates memory with causal abstractions for dynamic adaptation, while Generative Agents~\cite{park2023generative,chen2025hierarchical,chen2024gaussianvton,li2025text} retrieve experiences based on recency and relevance, forming structured reflections to support reasoning. MemGPT~\cite{packer2023memgpt} enables LLMs to manage working and long-term memory in task sequences. However, these approaches typically target static or narrow task domains, limiting their scalability in more diverse, interactive environments. \textit{Online learning} supports real-time adaptation~\cite{zhang2018adaptive}. Early works such as Reflexion~\cite{shinn2023reflexion}, AdaPlanner~\cite{sun2023adaplanner}, and ReAct~\cite{yao2023react} enable agents to adjust planning based on feedback. Reflexion~\cite{shinn2023reflexion} allows agents to reflect on failures, and RAP~\cite{kagaya2024rap} uses retrieved experiences for decision-making in task-switching. Voyager~\cite{wang2023voyager} lets agents store verified programs as reusable skills in dynamic environments. Despite progress, these methods often underperform in long-horizon tasks with delayed feedback and generally lack mechanisms for autonomous policy evolution. \textit{Limitations remain}: Most memory and online learning frameworks lack support for real-time, autonomous policy evolution in truly interactive, dynamic environments~\cite{feng2025evoagent}. AutoManual~\cite{chen2024automanual} introduces dynamic rules for adaptation in ALFWorld and MiniWoB++, while Agent-Pro~\cite{zhang2024agent} refines strategies in card games via policy-level reflection. However, both are limited to specific domains and lack generalizability. Furthermore, existing methods seldom allow for parameter-free, real-time adaptation post-deployment, restricting alignment with human-like continual learning~\cite{zhang2024cppo}. \textit{Our approach} addresses these gaps by proposing a human-inspired online learning workflow (see section~\ref{sec:methods}) that enables agents to abstract and update knowledge through continuous interaction—supporting dynamic, real-time, and parameter-free continual adaptation.

%% file: files/3-benchmark.tex
\section{Games in EvoEmpirBench}
\label{sec:Benchmark}

\begin{figure}[h]
\centering
\includegraphics[scale=0.55]{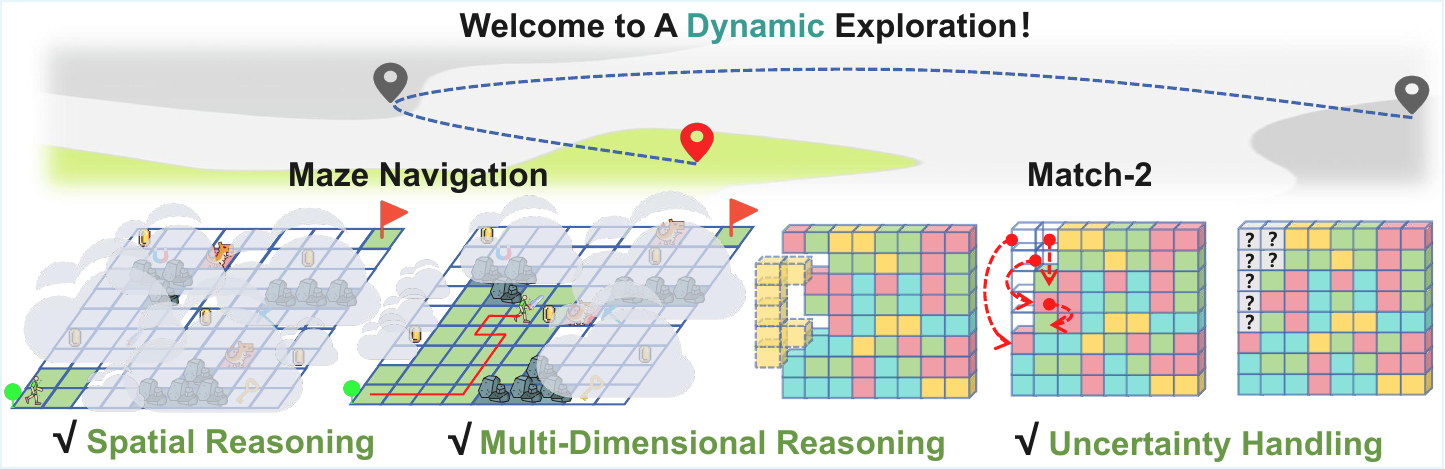}
\caption{Overview of EvoEmpirBench: Locally observable maze navigation (left) and match-2 elimination (right), illustrating dynamic and interactive challenges for language agents.}
\label{fig:benchmark_games}
\end{figure}

EvoEmpirBench (EEB) consists of two dynamic, interactive game environments designed to comprehensively evaluate LLM reasoning in complex, partially observable settings: a locally observable maze navigation task and a match-2 elimination game (Figure~\ref{fig:benchmark_games}). Both games simulate real-world uncertainty and agent-environment dynamics, demanding skills in spatial analysis, language understanding, and adaptive planning. Each game provides 3 progressively difficult levels (Easy, Medium, Hard) for both training and testing, totaling 120 diverse task episodes. EEB supports both human and agent operation, enabling flexible evaluation under different paradigms. We now outline the core rules, difficulty scaling, and evaluation metrics.

\subsection{Game Rules}
\label{subsec:game_rules}

\textbf{Maze Navigation.} In this 9$\times$9 grid world, agents act under local observability—exploring to reveal new tiles while aiming to reach a goal with high score. Main objectives include collecting five gold coins and exploring unknown areas efficiently. As difficulty increases, agents encounter new elements: Easy levels feature a simple map; Medium introduces two moving monsters that threaten agent health; Hard adds four interactive items—Pickaxe (breaks obstacles), Iron Sword (unlimited monster combat), Magnet (area coin collection), and Key (needed for exit)—along with the existing challenges. Many agent actions, such as obtaining weapons or destroying obstacles, alter the environment dynamically, requiring adaptive, multi-step reasoning.

\textbf{Match-2 Elimination.} Agents operate on an 8$\times$8 board, eliminating connected blocks of identical color (A–D) by matching two or more, with exponentially increasing rewards for larger matches. The primary goal is to meet per-color elimination targets within a limited number of moves, striving to maximize overall score and efficiency. Following each elimination, remaining blocks settle downward and new blocks spawn at the top. Agents may use step-limited props (row/col clear, bomb, hammer) at point costs to optimize gameplay. Difficulty levels vary step constraints and target requirements.

\subsection{Difficulty Settings}
\label{subsec:difficulty_settings}

As summarized in Table~\ref{tab:difficulty_settings}, EEB increases challenge by scaling environment complexity and resource constraints. In maze navigation, more complex maps, adversaries, and items are layered at higher levels to push planning and adaptation. In match-2, shrinking move allowances and rising elimination targets require increasingly efficient strategies.
\begin{table}[ht]
  \centering
  \caption{\textbf{Difficulty settings in EvoEmpirBench.} Maze navigation raises environmental complexity; match-2 tightens step budgets and goals.}
  \label{tab:difficulty_settings}
  \rowcolors{2}{gray!10}{white}
  \begin{adjustbox}{max width=\textwidth, center}
    \begin{tabular}{%
        C{2cm} C{2.4cm} C{2.6cm} C{3cm} |
        C{2cm} C{2.4cm} C{2.2cm} C{3cm}
      }
      \toprule
      \multicolumn{4}{c|}{\textbf{Maze Navigation}} & \multicolumn{4}{c}{\textbf{Match-2 Elimination}} \\
      \rowcolor{gray!30}
      \textbf{Level} & \textbf{Grid Size} & \textbf{Monsters} & \textbf{Items}
      & \textbf{Level} & \textbf{Grid Size} & \textbf{Steps} & \textbf{Target per Color} \\
      \midrule
      Easy   & 9$\times$9  & None       & 5 Coins               & Easy   & 8$\times$8  & 15--18 & 8--12  \\
      Medium & 9$\times$9  & 2 Moving   & 5 Coins               & Medium & 8$\times$8  & 12--15 & 12--16 \\
      Hard   & 9$\times$9  & 2 Moving   & 5 Coins + 4 Items     & Hard   & 8$\times$8  & 10--13 & 16--20 \\
      \bottomrule
    \end{tabular}
  \end{adjustbox}
\end{table}
\vspace{-0.5cm}

\subsection{Evaluation Metrics}
\label{subsec:evaluation_metrics}

Performance on each game is assessed with complementary metrics (Table~\ref{tab:evaluation_metrics}), reflecting both general task success and environment-specific skills. Key metrics include \textit{Suc.Rate} (Success Rate) and \textit{A.Score} (Average Score). For the maze environment, additional metrics capture navigation efficiency and survival capabilities: \textit{A.steps} (Average Steps), \textit{A.Explor} (Average Exploration), \textit{A.Gold} (Average Gold Collected), \textit{Rem.HP} (Remaining HP), \textit{A.kills} (Average Enemy Kills), and \textit{A.Barr} (Average Barriers Destroyed). In match-2, metrics emphasize step efficiency and API optimization: \textit{R/M.S} (Remaining/Max Steps Ratio), \textit{Score/Step}, \textit{Clear/Step}, and \textit{API Eff.} (API Efficiency). These metrics provides a thorough evaluation of agent capabilities across distinct task requirements.

\begin{table}[htbp]
\centering
\caption{\textbf{Evaluation metrics in EvoEmpirBench.}}
\label{tab:evaluation_metrics}
{\scriptsize
\rowcolors{3}{gray!10}{white} % 从第3行开始交替颜色
\begin{adjustbox}{max width=\textwidth, center}
\begin{tabular}{%
C{2.1cm} C{3.6cm} | C{2.1cm} C{3.6cm}% 微调列宽
}
\toprule
\multicolumn{2}{c|}{\textbf{Maze Navigation (8 metrics)}} & \multicolumn{2}{c}{\textbf{Match-2 Elimination (6 metrics)}} \\
\rowcolor{gray!30}
\bfseries Metric & \bfseries Description & \bfseries Metric & \bfseries Description \\
\midrule
Suc.Rate & Episode completion ratio & Suc.Rate & Episode completion ratio \\
A.Score & Mean cumulative reward & A.Score & Mean cumulative reward \\
A.steps & Average steps per episode & R/M.S & Remaining steps ratio \\
A.Explor & Explored map percentage & Score/Step & Reward per move \\
A.Gold & Collected gold ratio & Clear/Step & Blocks cleared per move \\
Rem.HP & Surviving health points & API Eff. & Valid API call ratio \\
A.kills & Average defeated enemies & \cellcolor{white} & \cellcolor{white} \\
A.Barr. & Average destroyed barriers & \multicolumn{2}{c}{\raisebox{-0.5ex}[0pt][0pt]{\textit{\footnotesize *Match-2 metrics complete here}}} \\
\bottomrule
\end{tabular}
\end{adjustbox}
}
\end{table}
\vspace{-0.3cm}

%% file: files/4-methodology.tex
\section{Methods}
\label{sec:methods}
\begin{figure}[t!]
    \centering
    \includegraphics[width=\textwidth]{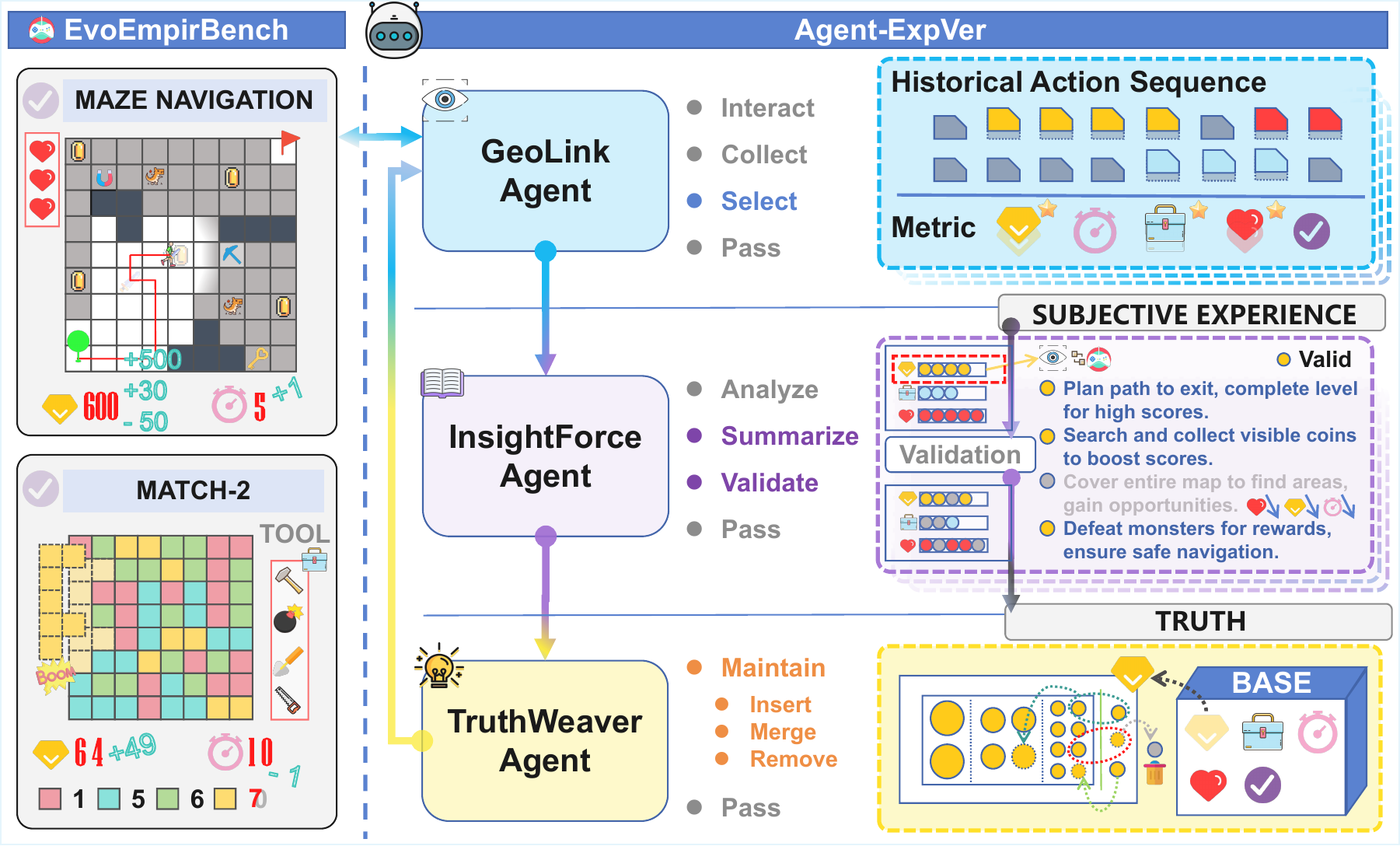}
    \caption{\textbf{Workflow of the Agent-ExpVer System.} The left side showcases EvoEmpirBench (EEB), our dynamic benchmark. The right side presents the Agent-ExpVer framework, comprising three agents: the GeoLink Agent collects and selects key historical actions based on game highlight metrics; the InsightForce Agent summarizes and validates subjective experiences; the TruthWeaver Agent maintains truths (insert, merge, remove) and passes them back to the GeoLink Agent. The figure highlights the processes of selection, summarization, validation, and maintenance.}
    \label{fig:method_workflow}
\end{figure}

To address the limitations of static training paradigms in dynamic, partially observable environments, we present a human-inspired online learning framework: \textbf{Agent-ExpVer} (\emph{Experience}+\emph{Verification}), with its workflow illustrated in Figure~\ref{fig:method_workflow}. In contrast to classic reinforcement learning~\cite{zhang2021learning,zhang2022closed}—which relies on large-scale pre-collected data and can struggle to generalize—Agent-ExpVer leverages in-context learning~\cite{dong2022survey} and continual experience abstraction~\cite{zhang2024agent}. The agent’s experiences are systematically organized, verified, and merged to build a reusable knowledge base that supports online adaptation.
\vspace{-0.4cm}
\subsection{Active Environment Interaction}
\label{subsec:env_interaction}

Human learners actively reduce environmental uncertainty through exploration. Agent-ExpVer mimics this with an environment interaction agent called \textit{GeoLink Agent} that iteratively interacts with the game environment.

At each timestep $t$, the agent observes the current state $\mathbf{s}_t$ (partially observable), chooses an action $a_t$ based on its policy $\pi_t$, and returns the subsequent state $\mathbf{s}_{t+1}$ and reward $r_t$:
\begin{align}
    a_t &\sim \pi_t(\mathbf{s}_t), \\
    (\mathbf{s}_{t+1}, r_t) &\sim \mathcal{E}(\mathbf{s}_t, a_t),
\end{align}
where $\mathcal{E}$ represents the environment dynamics. Across each episode, the agent accumulates an interaction history $\mathcal{H}_{0:T} = \{(\mathbf{s}_0,a_0,r_0), \ldots, (\mathbf{s}_T, a_T, r_T)\}$. This history is subsequently used for experience abstraction.

\subsection{Subjective Experience Abstraction and Truth Distillation}
\label{subsec:experience_abstraction}

After an episode, Agent-ExpVer employs a \emph{InsightForce agent}— inspired by human episodic memory—to process the collected trajectory $\mathcal{H}_{0:T}$ and final metrics $\mathbf{m}$ (A.Score, Suc.Rate, etc.).

The InsightForce agent summarizes key decisions and outcomes as a subjective experience $\mathbf{e}$:
\begin{equation}
    \mathbf{e} = f_\mathrm{sum}(\mathcal{H}_{0:T}, \mathbf{m})
\end{equation}
where $f_\mathrm{sum}$ is executed by prompting the LLM. Experiences are added to a memory module $\mathcal{M}_\mathrm{exp}$.

To verify and refine this knowledge, the agent then replays the episode, integrating $\mathbf{e}$ as a prompt extension to its policy. If the level is passed and resulting score improves, $\mathbf{e}$ is upgraded to a reusable \emph{truth}, stored in truth repository $\mathcal{M}_\mathrm{truth}$, managed by the \textit{TruthWeaver agent}:
\begin{equation}
    \mathcal{M}_\mathrm{truth} \leftarrow \mathcal{M}_\mathrm{truth} \cup \mathbf{e} \quad \text{if} \quad P \land (S' > S)
\end{equation}
where $P$ indicates whether the level is passed (\emph{true} if passed, \emph{false} otherwise), and $S' > S$ indicates an improved score.

% To avoid redundancy, truths with high semantic similarity (via cosine similarity in embedding space) are periodically merged into more general rules by LLM prompt:
% \begin{equation}
%     \text{sim}(\mathbf{e}_i, \mathbf{e}_j) = \frac{\mathbf{v}_i \cdot \mathbf{v}_j}{\|\mathbf{v}_i\| \|\mathbf{v}_j\|}
% \end{equation}
% where $\mathbf{v}_i,\mathbf{v}_j$ are LLM embeddings of $\mathbf{e}_i$ and $\mathbf{e}_j$ respectively.
The TruthWeaver Agent manages the incoming truths through the following operations: (1) merging incoming truths that have high semantic similarity with existing truths, (2) removing incoming truths that are redundant with existing truths, and (3) inserting new truths.

\subsection{Policy Evolution via Truth Integration}
\label{subsec:policy_evolution}

The current agent policy is constructed by combining the base prompt $\pi_0$ with all truths $\mathcal{M}_\mathrm{truth}$:
\begin{equation}
    \pi_t = \pi_0 \cup \bigcup_{\mathbf{e} \in \mathcal{M}_\mathrm{truth}} \mathbf{e}
\end{equation}
After each episode, new truths are added and validated as above, enabling continual improvement as the system accumulates and distills knowledge.

We define policy improvement as the average test score increment:
\begin{equation}
    \Delta = \frac{1}{N} \sum_{i=1}^N (S_i^t - S_i^{t-1}),
\end{equation}
where $N$ is the number of test cases, $S_i^t$ and $S_i^{t-1}$ are scores using the current and previous policies, respectively. If policy performance degrades ($\Delta < 0$), the newest updates are reverted and abstraction is repeated.

\vspace{1ex}
\noindent
This workflow supports continual, online learning, allowing Agent-ExpVer to bridge the gap between static benchmark settings and the adaptive problem-solving required in dynamic, real-world scenarios.

%% file: files/5-experiment.tex
\section{Experiment}
\label{sec:experiment}
\textbf{Experiment Setting.} We evaluated both proprietary and open-source models on our EvoEmpirBench (EEB). Proprietary models include GPT-4~\cite{achiam2023gpt}, GPT-4.1, Gemini-2.0-flash, Gemini-2.5-flash-preview~\cite{leng2025crosswordbenchevaluatingreasoningcapabilities}, Claude-3-5-sonnet~\cite{anthropic_claude_2024}, Claude-3-7-sonnet, and Grok-3. Open-source models comprise Deepseek-V3~\cite{liu2024deepseek}, Llama-3.1-8B-instruct, Llama-3.1-70B-instruct~\cite{dubey2024llama}, Qwen2.5-7B-instruct, Qwen2.5-14B-instruct, Qwen2.5-32B-instruct, Qwen2.5-72B-instruct, and Qwen3-30B-a3B~\cite{qwq32b}. EEB features two tasks—maze navigation and match-2 games—each at three difficulty levels with 30 instances per level (90 instances per task). We also applied our Agent-ExpVer workflow to GPT-4.1, Gemini-2.5-flash-preview, Claude-3-7-sonnet, and Qwen2.5-32B-instruct, observing marked improvements in their EEB performance. Finally, by benchmarking both models and human participants, we provide a clear view of current LLM capabilities and limitations in dynamic reasoning.

% \MW{to do}\textbf{Experiment Setting.} In this study, we systematically evaluate a range of state-of-the-art proprietary models—including GPT-4~\cite{achiam2023gpt}, GPT-4.1~\cite{}, Gemini-2.0-flash~\cite{}, Gemini-2.5-flash-preview~\cite{}, Claude-3-5-sonnet~\cite{}, Claude-3-7-sonnet~\cite{}, and Grok-3~\cite{}—as well as leading open-source models such as Deepseek-V3~\cite{liu2024deepseek}, Llama-3.1-8b-instruct~\cite{}, Llama-3.1-70b-instruct~\cite{}, Qwen2.5-7b-instruct~\cite{}, Qwen2.5-14b-instruct~\cite{}, Qwen2.5-32b-instruct~\cite{}, Qwen2.5-72b-instruct~\cite{}, and Qwen3-30b-a3b~\cite{} on EvoEmpirBench(EEB), which is designed to assess model reasoning on complex tasks. EEB consists of two main tasks—maze navigation and match-2 games—each with three difficulty levels and 30 samples per level, resulting in 90 unique instances per task. Furthermore, we applied our Agent-ExpVer workflow to selected models, including GPT-4.1, Gemini-2.5-flash-preview, Claude-3-7-sonnet, and Qwen2.5-32b-instruct, and observed that this workflow significantly improved their reasoning abilities on EEB. By comparing the results of both models and humans, we are able to more accurately benchmark the progress and limitations of current LLMs in complex dynamic reasoning scenarios.

\begin{table}[!ht]
  \centering
  \caption{
    \textbf{“All Levels” Metrics Across Models — Maze Navigation Game Performance.}
Metric abbreviations:
Suc.Rate (Success Rate), A.Score (Avg. Score), A.Steps (Avg. Steps), A.Explor. (Avg. Exploration), A.Gold (Gold Collection Rate), Rem.HP (Remaining HP), A.Kills (Avg. Kills), A.Barr. (Avg. Barrier Interactions).
Top-3 values in each column are highlighted in \textcolor{gold}{gold}, \textcolor{silver}{silver}, and \textcolor{bronze}{bronze}.}
  \label{tab:all_levels_maze_highlight}

  \scriptsize
  \setlength{\tabcolsep}{4pt}
  \renewcommand{\arraystretch}{1.1}
  \rowcolors{2}{gray!10}{white}
  \begin{adjustbox}{max width=\textwidth,center}
    \begin{tabular}{>{\raggedright}p{4.5cm} c c c c c c c c c}
      \toprule
      \rowcolor{gray!30}
      \textbf{Model} & \textbf{Samp.} & \textbf{Suc.Rate\,↑} & \textbf{A.Score\,↑} & \textbf{A.steps\,↓} & \textbf{A.Explor.\,↑} & \textbf{A.Gold\,↑} & \textbf{Rem.HP\,↑} & \textbf{A.kills\,↑} & \textbf{A.Barr.\,↑} \\
      \midrule
      
      \multicolumn{10}{l}{\textbf{Human Baseline}} \\
      human & 90 & 90.00 & 2914.67 & 20.6 & 77.54 & 68.22 & 2.43 & 0.83 & 1.43 \\
      
      \multicolumn{10}{l}{\textbf{Open-Source Models}} \\
      \midrule
      deepseek-V3 & 90 & \cellcolor{silver!30}61.11 & \cellcolor{silver!30}1649.78 & 50.61 & \cellcolor{silver!30}83.15 & \cellcolor{gold!30}75.56 & \cellcolor{silver!30}1.71 & \cellcolor{gold!30}0.77 & \cellcolor{gold!30}0.80 \\
      llama-3.1-8b-instruct & 90 & 23.33 & -1213.67 & 54.42 & 51.00 & 24.89 & 1.28 & 0.03 & 0.27 \\
      llama-3.1-70b-instruct & 90 & 31.11 & 489.89 & \cellcolor{bronze!30}39.19 & 58.30 & 51.11 & 0.87 & \cellcolor{bronze!30}0.30 & 0.23 \\
      qwen2.5-7b-instruct & 90 & 3.33 & -1071.67 & 62.74 & 65.45 & 51.56 & 0.53 & 0.27 & \cellcolor{bronze!30}0.60 \\
      qwen2.5-14b-instruct & 90 & 36.67 & 24.67 & 53.69 & 65.31 & 55.56 & 1.30 & 0.27 & 0.23 \\
      qwen2.5-32b-instruct & 90 & 42.22 & 1122.22 & 38.41 & 68.43 & 60.44 & 1.12 & 0.23 & 0.50 \\
      \textbf{qwen2.5-32b-instruct-ours} & 90 & \cellcolor{bronze!30}54.44 & \cellcolor{bronze!30}1532.33 & \cellcolor{silver!30}38.81 & 68.73 & 61.33 & 1.18 & 0.26 & 0.5 \\
      qwen2.5-72b-instruct & 90 & 34.44 & 368.67 & 54.76 & \cellcolor{bronze!30}75.11 & \cellcolor{bronze!30}67.33 & \cellcolor{bronze!30}1.34 & \cellcolor{bronze!30}0.30 & 0.43 \\
      qwen3-30b-a3b & 90 & \cellcolor{gold!30}65.56 & \cellcolor{gold!30}2359.22 & \cellcolor{gold!30}33.10 & \cellcolor{gold!30}81.93 & \cellcolor{silver!30}73.33 & \cellcolor{gold!30}1.73 & \cellcolor{silver!30}0.43 & \cellcolor{bronze!30}0.57 \\
      
      \multicolumn{10}{l}{\textbf{Proprietary Models}} \\
      \midrule
      grok-3 & 90 & 67.78 & \cellcolor{gold!30}2893.89 & \cellcolor{silver!30}29.66 & \cellcolor{bronze!30}83.47 & 72.67 & 1.70 & \cellcolor{gold!30}0.93 & \cellcolor{gold!30}1.03 \\
      gemini-2.0-flash & 90 & 28.89 & -627.44 & 67.51 & 65.42 & 60.22 & 1.54 & 0.30 & 0.43 \\
      gemini-2.5-flash-preview & 90 & 45.56 & 1426.11 & 48.30 & 83.37 & \cellcolor{bronze!30}74.44 & 1.14 & \cellcolor{silver!30}0.77 & \cellcolor{bronze!30}0.80 \\
      \textbf{gemini-2.5-flash-preview-ours} & 90 & 64.44 & 2158.67 & 36.88 & 80.81 & 71.33 & 1.68 & 0.50 & 0.70 \\
      claude-3-5-sonnet & 90 & 50.00 & 1664.00 & 42.44 & 81.70 & 73.56 & 1.34 & 0.57 & 0.67 \\
      claude-3-7-sonnet & 90 & 68.89 & 2793.44 & \cellcolor{gold!30}27.40 & 82.59 & 73.33 & 1.70 & 0.47 & 0.77 \\
      \textbf{claude-3-7-sonnet-ours} & 90 & \cellcolor{bronze!30}72.22 & 2858 & \cellcolor{bronze!30}31.33 & 83.35 & \cellcolor{gold!30}76.44 & \cellcolor{bronze!30}1.78 & 0.73 & 0.83 \\
      gpt-4 & 90 & 43.33 & 1623.33 & 32.59 & 73.11 & 64.44 & 1.12 & 0.33 & 0.57 \\
      gpt-4.1 & 90 & 73.33 & 2562.33 & 34.03 & 83.63 & \cellcolor{silver!30}74.89 & \cellcolor{silver!30}1.98 & 0.47 & 0.67 \\
      \textbf{gpt-4.1-ours}  & 90 & \cellcolor{gold!30}78.89 & \cellcolor{bronze!30}2805.67 & 32.77 & \cellcolor{gold!30}83.69 & 72.67 & \cellcolor{gold!30}1.99 & \cellcolor{bronze!30}0.73 & 0.77 \\
      \bottomrule
    \end{tabular}
  \end{adjustbox}
  \vspace{-2em}
\end{table}
\textbf{Maze Navigation Game.} As shown in Table~\ref{tab:all_levels_maze_highlight}, applying \textbf{Agent-ExpVer} to both open-source and proprietary models boosts their Maze Navigation performance by an average of \(\,+5.6\%\) in success rate and \(\,+9.5\%\) in mean score. In particular, \textbf{GPT-4.1} achieves a \(\,+5.56\%\) (absolute) / \(\,+7.6\%\) (relative) increase in success and a \(\,+243.3\)–point / \(\,+9.5\%\) rise in score, making it the closest to human-level performance. Among open-source models, \textbf{Qwen-32B-instruct} with Agent-ExpVer posts \(\,+12.22\%\) / \(\,+28.96\%\) success and \(\,+410.1\)–point / \(\,+36.6\%\) score gains, surpassing nearly all peers and rivaling Deepseek-V3 and Qwen-30B-A3B.

% Beyond aggregate metrics, stronger agents complete mazes with fewer steps, retain higher health, collect more coins, eliminate more monsters, and clear more obstacles—reflecting improved spatial reasoning, risk–reward balancing, and long-term planning. 

% \textbf{Maze Navigation Game} As illustrated in Table~\ref{tab:all_levels_maze_highlight}, current mainstream open-source and proprietary models demonstrate varying performance in the Maze Game task of EvoEmpirBench. Our analysis reveals that models enhanced through the Agent-ExpVer workflow exhibit substantial improvements over their base counterparts, achieving average increases of 5.6\% in success rate and 9.5\% in mean score. Notably, the GPT-4.1 model attained a (+5.56\% absolute, +7.6\% relative) enhancement in success rate and (+243.3 points, +9.5\% relative) improvement in average score, positioning it as the LLM closest to the human baseline among evaluated systems.

% For open-source alternatives, the Agent-ExpVer optimized qwen-32b-instruct model shows competitive performance with (+12.22\% absolute, +28.96\% relative) success rate gain and (+410.1 points, +36.6\% relative) score improvement. This advancement enables it to surpass most open-source counterparts while approaching the performance levels of deepseek-V3 and qwen-30b-a3b models.

The comprehensive metrics in EvoEmpirBench effectively evaluate model reasoning capabilities in complex environments. Our analysis reveals that stronger models achieve higher completion rates with fewer steps, while maintaining greater residual health points and demonstrating superior performance in coin collection, monster elimination, and obstacle destruction. These metrics collectively assess spatial reasoning in dynamic environments, risk-reward optimization, long-term versus short-term reward trade-offs, and interactive proficiency with virtual ecosystems. 

%As demonstrated in Figure 6, models equipped with Agent-ExpVer's verified knowledge base exhibit enhanced survival strategies and optimized pathfinding efficiency, effectively mitigating potential risks and step penalties. Through iterative interactions with EvoEmpirBench's sophisticated environment, agents progressively develop strategic cognition - a learning process that enables effective long-term planning and balanced decision-making between environmental challenges and reward acquisition.

% \begin{figure}[h]
%     \centering
%     % 左图 minipage（占50%宽度，可调整比例）
%     \begin{minipage}[t]{0.36\textwidth} % 0.48避免左右间距过宽
%         \centering
%         \includegraphics[width=\textwidth]{files/fig/maze_game_single_radar_bold.pdf} % 宽度占满minipage
%         \caption{Model performance in Maze Game}
%         \label{fig:Maze_Game_performance}
%     \end{minipage}% 注意%符号避免换行间距
%     \hfill % 水平填充间距
%     % 右图 minipage（与左图参数一致）
%     \begin{minipage}[t]{0.62\textwidth}
%         \centering
%         \includegraphics[width=\textwidth]{files/fig/maze_game_levels.pdf}
%         \caption{The Difficulty Distribution Of Match-2 Game}
%         \label{fig:truth_learning}
%     \end{minipage}
%     \caption{Maze Game}
%     \label{fig:double}
% \end{figure}

%----------------------------
% Table 2: Match-2 Puzzle
%----------------------------

\begin{table}[!ht]
  \centering
  \caption{%
\textbf{“All Levels” Metrics Across Models — Match-2 Elimination Game Performance.}
Top-3 values (exclude the
human baseline) in each category are highlighted in \textcolor{gold}{gold}, \textcolor{silver}{silver}, and \textcolor{bronze}{bronze}.
Metric abbreviations: Suc.Rate (Success Rate), A.Score (Avg. Score), R/M.S (Redundant/Missing Steps), Score/Step (Score per Step), Clear/Step (Clearance per Step), API Eff. (API Efficienc)}
  \label{tab:Match-2-all-levels}
  
  {  \scriptsize
  \setlength{\tabcolsep}{5pt}
  \renewcommand{\arraystretch}{0.9}
  \rowcolors{2}{gray!10}{white}
  \begin{adjustbox}{max width=\textwidth,center}
    \begin{tabular}{%
      >{\raggedright}C{4.2cm}
      C{0.9cm}
      C{1.2cm}
      C{1.5cm}
      C{1.4cm}
      C{1.4cm}
      C{1.4cm}
      C{1.4cm}
    }
      \toprule
      \rowcolor{gray!30}
      \textbf{Model} & \textbf{Sample} & \textbf{Suc.Rate\,↑} & \textbf{A.score\,↑} & \textbf{R/M.S\,↑} & \textbf{Score/Step\,↑} & \textbf{Clear/Step\,↑} & \textbf{API Eff.\,↑} \\
      \midrule

      \multicolumn{8}{l}{\textbf{Human Baseline}} \\
      human & 90 & 86.67 & 350.2 & 22.99 & 34.40 & 6.20 & \,-- \\

      \multicolumn{8}{l}{\textbf{Open-Source Models}} \\
      \midrule
      deepseek-V3 & 90 & \cellcolor{silver!30}37.78 & \cellcolor{bronze!30}218.29 & \cellcolor{silver!30}8.99 & \cellcolor{bronze!30}17.07 & \cellcolor{bronze!30}4.92 & \cellcolor{bronze!30}44.55 \\
      llama-3.1-8b-instruct & 90 & 22.22 & 95.4 & \cellcolor{silver!30}4.18 & 9.05 & 4.33 & 4.95 \\
      llama-3.1-70b-instruct & 90 & 30.00 & \cellcolor{gold!30}289.67 & 4.94 & \cellcolor{gold!30}21.59 & 4.65 & 32.82 \\
      qwen2.5-7b-instruct & 90 & 17.78 & 85.2 & 3.61 & 8.09 & 3.42 & 2.19 \\
      qwen2.5-14b-instruct & 90 & 36.67 & 170.91 & \cellcolor{gold!30}9.38 & 13.42 & 4.86 & 37.86 \\
      qwen2.5-32b-instruct & 90 & 33.33 & 203.07 & \cellcolor{bronze!30}8.80 & 15.60 & 5.11 & 34.29 \\
      \textbf{qwen2.5-32b-instruct-ours} & 90 & \cellcolor{gold!30}41.11 & 197.42 & 8.93 & 15.52 & \cellcolor{gold!30}5.30 & \cellcolor{gold!30}57.76 \\
      qwen2.5-72b-instruct & 90 & 35.56 & 194.36 & 7.06 & 14.86 & 4.88 & 38.87 \\
      qwen3-30b-a3b & 90 & 38.89 & \cellcolor{silver!30}266.30 & 6.43 & \cellcolor{silver!30}20.31 & 4.91 & \cellcolor{silver!30}54.43 \\
      
      \multicolumn{8}{l}{\textbf{Proprietary Models}} \\
      \midrule
      grok-3 & 90 & \cellcolor{bronze!30}42.22 & 201.11 & \cellcolor{bronze!30}11.04 & 15.87 & 5.08 & 35.43 \\
      gemini-2.0-flash & 90 & 34.44 & 202.24 & 8.52 & 15.76 & \cellcolor{bronze!30}5.12 & 49.19 \\
      gemini-2.5-flash-preview & 90 & 37.78 & \cellcolor{silver!30}410.76 & \cellcolor{bronze!30}5.57 & \cellcolor{silver!30}30.58 & 4.85 & 69.73 \\
      \textbf{gemini-2.5-flash-preview-ours} & 90 & 37.78 & \cellcolor{gold!30}415.49 & 7.29 & \cellcolor{gold!30}32.33 & 5.04 & \cellcolor{gold!30}88.85 \\
      claude-3-5-sonnet & 90 & 26.67 & \cellcolor{bronze!30}342.02 & 4.88 & 25.26 & 4.57 & 53.27 \\
      claude-3-7-sonnet & 90 & 41.11 & 298.96 & 7.76 & 23.04 & 4.92 & 50.49 \\
      \textbf{claude-3-7-sonnet-ours} & 90 & \cellcolor{silver!30}47.19 & 291.01 & \cellcolor{gold!30}17.49 & 26.70 & \cellcolor{silver!30}5.64 & \cellcolor{silver!30}88.06 \\
      gpt-4 & 90 & 31.11 & 142.27 & 5.26 & \cellcolor{bronze!30}27.80 & 4.43 & 36.67 \\
      gpt-4.1 & 90 & 40.00 & 245.04 & 7.52 & 18.67 & 5.03 & 46.12 \\
      \textbf{gpt-4.1-ours} & 90 & \cellcolor{gold!30}53.33 & 234.6 & \cellcolor{silver!30}17.31 & 19.90 & \cellcolor{gold!30}5.86 & \cellcolor{bronze!30}75.72 \\
      \bottomrule
    \end{tabular}
  \end{adjustbox}
  }
\end{table}

\textbf{Match-2 Elimination Game.}
Table~\ref{tab:Match-2-all-levels} highlights the greater challenge of the Match-2 task, where baseline LLMs average only \(\,33.7\%\) success. Open-source contenders Deepseek-V3 (37.78\%) and Qwen-30B-a3b (38.89\%), and proprietary models Grok-3 (42.22\%), Claude-3-7-sonnet (41.11\%), and GPT-4.1 (40.00\%), all underperform humans by a wide margin. Integrating Agent-ExpVer yields consistent gains of \(\,+13.3\%\) in success rate, \(\,+9.8\%\) more steps remaining, \(\,+1.2\) elimination score per step, and \(\,+0.8\) eliminations per step.

Notably, GPT-4.1 achieves \(\,+13.33\%\) success—becoming the top non-human agent—though its average score dips by 4.3\% (–10.44 points). Conversely, Gemini-2.5-flash-preview maintains its success rate while gaining \(\,+1.2\%\) in score (+4.73 points), even surpassing human benchmarks. This underscores EEB’s emphasis on compound objectives: agents must judiciously trade immediate elimination efficiency (via in-game items like bombs) against long-term score maximization.

\subsection{Ablation Studies}
\label{ssec:ablations}

\textbf{Is organizing and managing a truth repository important?}
To assess the importance of managing the truth repository, we conduct experiments on the EEB benchmark, comparing the full Agent-ExpVer framework against a variant without the TruthWeaver Agent, using Qwen2.5-32B and GPT-4.1 as bases. As reported in Table~\ref{tab:cross_game_ablation}, removing TruthWeaver causes a marked drop in both tasks: for Qwen2.5-32B, average scores fall by 16.9\% and success rates decline by 6.1\%. Moreover, reasoning efficiency suffers—maze completion requires 3.5\% more steps, and remaining steps in Match-2 decrease by 4.0\%. For GPT-4.1, average scores fall by 5.8\% and success rates decline by 8.3\%; maze completion requires 0.9\% more steps, while remaining steps in Match-2 decrease by 26.8\%. These results confirm that TruthWeaver is essential for (1) mitigating catastrophic forgetting via dynamic truth consolidation and (2) stabilizing belief distributions to avoid redundant hypotheses.

% \textbf{Is organizing and managing a truth repository important?} To validate the critical role of knowledge base management, we conduct an ablation study on the TruthWeaver Agent module by comparing the complete Agent-ExpVer framework against its variant without this component, using Qwen2.5-32B and GPT-4.1 as baseline models. As shown in Table 6, removing the TruthWeaver Agent leads to significant performance degradation across both tasks. The ablated framework exhibits an average score reduction of ?\% relative to baselines, accompanied by a ?\% decline in success rates. Furthermore, the modified version demonstrates compromised reasoning efficiency: it requires ?\% more steps to complete maze navigation tasks and achieves ?\% fewer remaining steps in Match-2 games, indicating diminished long-term planning capability.

% These results quantitatively confirm the module's effectiveness in addressing two fundamental challenges: dynamic mitigation of catastrophic forgetting during extended contextual interactions through structured knowledge consolidation, and stabilization of truth-weight distributions by preventing redundant hypothesis adoption.

% 消融实验1 表格
\begin{table}[!ht]
  \centering
\caption{\textbf{Cross‑Game Ablation Study of the TruthWeaver Framework.} Metrics for the \textit{Maze Game} (Cols 3–5) and \textit{Match‑2 Game} (Cols 6–8).}
  \label{tab:cross_game_ablation}
  \scriptsize
  \setlength{\tabcolsep}{7pt}
  \renewcommand{\arraystretch}{0.7}
  \rowcolors{2}{gray!10}{white}
  \begin{tabular}{l
                  r
                  >{\centering}p{1cm}
                  >{\centering}p{1cm}
                  >{\centering}p{1cm}
                  >{\centering}p{1cm}
                  >{\centering}p{1cm}
                  >{\centering\arraybackslash}p{1cm}}
    \toprule
    \rowcolor{gray!20}
    & & \multicolumn{3}{c}{\textbf{Maze Game}} & \multicolumn{3}{c}{\textbf{Match-2 Game}} \\
    \cmidrule(lr){3-5} \cmidrule(lr){6-8}
    \rowcolor{gray!30}
    \textbf{Model} & \textbf{Samples} &
    \textbf{Suc.Rate~↑} & \textbf{A.Score~↑} & \textbf{A.Steps~↓} &
    \textbf{Suc.Rate~↑} & \textbf{A.Score~↑} & \textbf{R/M.S.~↓} \\
    \midrule
    gpt-4.1                      & 90 & 73.33 & 2562.33 & 34.03 & 40.00 & 245.04 & 7.52  \\
    gpt-4.1 W/o.TW               & 90 & 77.78 & 2765.33 & 33.06 & 48.89 & 219.53 & 12.66 \\
    \textbf{gpt-4.1-ours}         
    & 90 & 78.89 & 2805.67 & 32.77 & 53.33 & 234.60 & 17.31 \\
    qwen2.5-32b-instruct         & 90 & 42.22 & 1122.22 & 38.41 & 33.33 & 203.07 & 8.80  \\
    qwen2.5-32b-instruct W/o.TW  & 90 & 51.11 & 1272.11 & 37.07 & 35.56 & 187.89 & 8.57  \\
    \textbf{qwen2.5-32b-instruct-ours}   & 90 & 54.44 & 1532.33 & 35.81 & 41.57 & 197.42 & 8.93  \\

    \bottomrule
  \end{tabular}
\end{table}

% \textbf{Why is EEB more challenging for model reasoning than existing benchmarks?} To better align with real-world characteristics of dynamic information acquisition and diverse interaction mechanisms, EEB introduces two critical designs: (1) In Maze in EEEB, agents operate under partial visibility conditions, requiring progressive environmental exploration to obtain complete information; (2) In Match-2 in EEB, multiple consumable tools (e.g., bombs facilitating level completion) can be purchased through point consumption. This design necessitates agents to make strategic decisions regarding optimal timing for tool usage while balancing short-term gains against long-term benefits.

\textbf{How much do EEB’s constraints affect model performance?} To evaluate, we conduct ablation studies on both games. As reported in Table~\ref{tab:Abbreviations2}, in the Maze Game, removing exploration via global visibility (Full-Vision) leads to a notable performance boost: GPT-4.1’s success rate rises from 73.33\% (EEB) to 93.33\%, and its average score increases from 2562.33 to 3412.22. Qwen2.5 shows a similar trend, with success rate improving from 42.22\% to 57.78\%. Both models also require fewer steps under Full-Vision, indicating reduced reasoning complexity.In the Match-2 Game, prohibiting tool use (NoProps) slightly improves scores for both models, but the gains are less pronounced than in Maze. For example, GPT-4.1’s success rate increases from 40.00\% (EEB) to 43.33\%, and Qwen2.5’s score rises from 203.07 to 411.62.These results demonstrate that EEB’s design—through partial observability and tool-based reasoning—significantly increases task difficulty and reasoning demands, as evidenced by consistent performance drops compared to control variants.

%消融实验2 表格
\vspace{-1mm}
\begin{table}[ht]
\centering
\caption{\textbf{Model Performance Comparison on Maze and Match‑2 Games.} Explanation of Metrics and Abbreviations:
EEB = Embodied Environment Baseline. 
Full-Vision = Model has full visual information. 
NoProps = No propositional information provided.}
\label{tab:Abbreviations2}
\scriptsize
\setlength{\tabcolsep}{8pt}
\renewcommand{\arraystretch}{0.7}
\rowcolors{2}{gray!10}{white}
\begin{tabular}{l *{8}{c}}
\toprule
\textbf{Metric} & 
\multicolumn{2}{c}{\textbf{GPT-4.1 (Maze)}} & 
\multicolumn{2}{c}{\textbf{Qwen2.5 (Maze)}} & 
\multicolumn{2}{c}{\textbf{GPT-4.1 (Match-2)}} & 
\multicolumn{2}{c}{\textbf{Qwen2.5 (Match-2)}} \\
\cmidrule(lr){2-3} \cmidrule(lr){4-5} \cmidrule(lr){6-7} \cmidrule(lr){8-9}
& \textbf{EEB} & \textbf{Full‑Vision} & \textbf{EEB} & \textbf{Full‑Vision} & \textbf{EEB} & \textbf{NoProps} & \textbf{EEB} & \textbf{NoProps} \\
\midrule
\textbf{Sample} &
90 & 90 & 90 & 90 & 90 & 90 & 90 & 90 \\
\textbf{Ave.Succ (\%)~↑} &
73.33 & 93.33 &
42.22 & 57.78 &
40.00 & 43.33 & 33.33 & 32.22 \\
\textbf{Ave.Score~↑} &
2562.33 & 3412.22 &
1122.22 & 1737.78 &
245.04 & 412.67 & 203.07 & 411.62 \\
\textbf{Ave.Steps~↓} &
34.03 & 19.53 &
38.41 & 18.47 &
7.52 & 10.04 & 8.80 & 6.24 \\
\bottomrule
\end{tabular}
% \vspace{-1em}
\end{table}

\subsection{Further Analysis}
\vspace{-1mm}
\textbf{What truth did the model learn with Agent-ExpVer?} With Agent-ExpVer, models like \textit{gpt-4.1-ours} and \textit{claude-3-7-sonnet-ours} in the Maze Game showed greater risk awareness—achieving higher remaining health points (1.99 and 1.78) but slightly reduced exploration and gold collection (Figure 6(a)). This suggests they internalized that survival is key for long-term success, sometimes favoring caution over immediate rewards. At the same time, increases in average kills and barrier interactions indicate more proactive engagement. In Match-2, models such as \textit{gpt-4.1-ours} and \textit{gemini-2.5-flash-preview-ours} reached higher success rates (53.33\% and 41.11\%) post-training, though sometimes with lower average scores. This trade-off reflects a shift toward strategies that prioritize completion, supported by gains in clearance per step and API efficiency, and demonstrates more efficient, deliberate actions.

\begin{figure}[!t]
\centering
\hspace*{-0.5cm} % 向左移动 2 厘米
\includegraphics[scale=0.45]{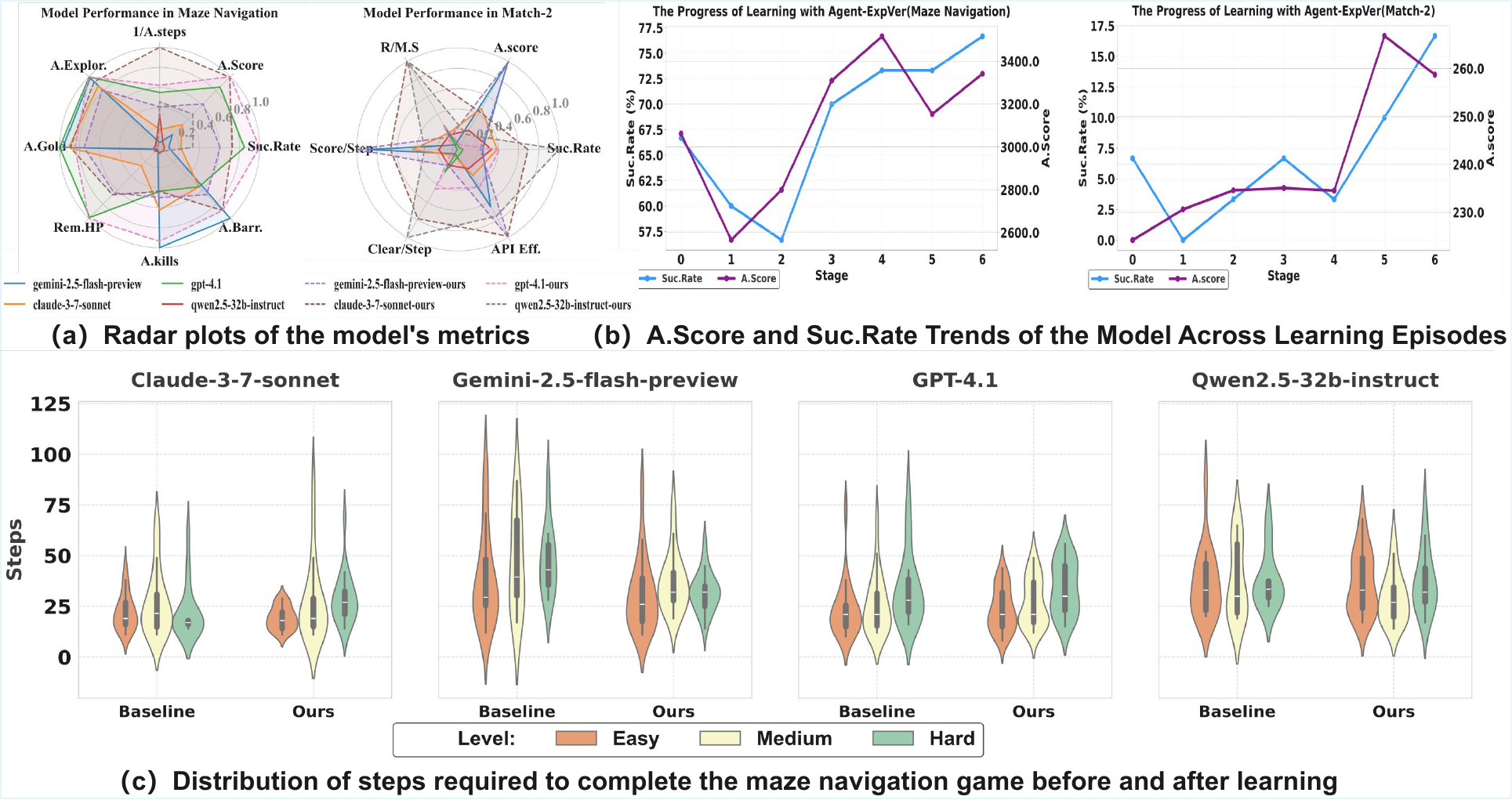}
\vspace{-1.8em}
\caption{~\textbf{Performance Analysis of Agent-ExpVer in EEB.} Demonstrating Enhanced Global Long-Horizon Reasoning through Dynamic Truth Refinement: (a) Radar Plots of Model Metrics, (b) A.Score and Suc.Rate Trends Across Learning Episodes, (c) Distribution of Steps for Maze Navigation Before and After Learning.}
\label{fig:further_analysis}
\vspace{-1em}
\end{figure}

The evolution of these “truths” is further illustrated in Figure~\ref{fig:further_analysis}(B). In the early rounds of truth induction, both Maze and Match-2 models experienced a temporary drop in success rate and score (e.g., Maze success rate fell from 66.67\% to 56.67\%, and Match-2 from 6.67\% to 0\%), likely due to the immaturity and overgeneralization of initial hypotheses. For instance, a model might infer that “bold exploration yields high rewards” after a lucky episode, only to encounter negative outcomes when this risky strategy is applied indiscriminately. Notably, the improvement observed in the Maze task can be attributed to the model’s ability to learn survival-oriented principles through the ExpVer module—details of which are provided in Appendix A. However, after 3–4 rounds of iterative learning, both success rate and score steadily improved (Maze: success rate rose to 76.66\%, score to 3370; Match-2: success rate to 16.67\%, score to 258), reflecting the model’s ability to refine its truths—shifting from reckless exploration to more cautious, stepwise progress.

Besides, Figure~\ref{fig:further_analysis}(c) reveals that models trained with Agent-ExpVer not only improved in aggregate metrics but also became more efficient in their execution. The distribution of steps required for successful completion narrowed and shifted toward lower values for both \textit{gpt-4.1-ours} and \textit{gemini-2.5-flash-preview-ours}, indicating enhanced planning and decision-making efficiency. In Maze, this translates to faster, safer navigation with reduced exposure to hazards, while in Match-2, it results in higher clearance rates and scores per step.

\label{ssec:discussion}

%% file: files/6-conclusion.tex
\section{Conclusion}
\label{sec:conclusion}
\vspace{-1mm}
We introduce EvoEmpirBench, a benchmark for spatial and high‑level reasoning in dynamic, interactive environments, featuring Maze Navigation and Match‑2 tasks. We also present Agent‑ExpVer, a three‑agent framework for environment interaction, experience synthesis, and adaptive truth management; experiments show it drives effective online learning and markedly improves agent reasoning and interactivity. \textbf{Limitations:} Performance remains tied to model capacity—smaller models lag, and even top systems fall short of human baselines. \textbf{Future Work:} We will boost Agent‑ExpVer’s adaptability (especially for lightweight models), expand EvoEmpirBench with tasks on temporal reasoning and multi‑agent collaboration, and develop advanced mechanisms for truth induction and experience management.

% In this work, we present EvoEmpirBench, a benchmark designed to rigorously assess models’ spatial and high-level reasoning in dynamic, interactive settings. EvoEmpirBench includes two tasks: Maze Navigation, which features partial observability, dynamic monsters, and interactive items, and Match-2, which challenges agents to strategically use tools and balance short- and long-term rewards. We also introduce Agent-ExpVer, a human-inspired framework with three agents responsible for environment interaction, experience synthesis, and dynamic truth management. Experiments show that Agent-ExpVer enables effective online learning, and we observe clear improvements in models’ reasoning and interactive capabilities within complex environments.

% \textbf{Limitation}

% Despite these advances, our approach is still constrained by the underlying model capacity. Models with smaller parameter counts continue to underperform even after Agent-ExpVer training, and even state-of-the-art models still exhibit a noticeable gap compared to human baselines.

% \textbf{Future Work}

% Future research will focus on enhancing the adaptability and generalization of Agent-ExpVer, particularly for smaller models and more diverse environments. We also plan to extend EvoEmpirBench with additional tasks that further challenge temporal reasoning and multi-agent collaboration, and to explore more sophisticated mechanisms for truth induction and experience management within the Agent-ExpVer framework.

%% file: files/appendix.tex
\appendix

\appendix
\section*{Appendix A: ExpVer-Agent Mechanism for Generation-Oriented Truth Formation}

This appendix details the ExpVer-Agent mechanism, an online learning approach based on ``subjective experience--verification--truth induction,'' designed to enhance agent performance in maze navigation tasks. The core procedure consists of the following steps:

\textbf{Generation of Subjective Experience}

In each level, the agent first acts as an ``explorer,'' completing a full game episode. During this process, the system automatically records the agent’s reasoning (i.e., decision-making analysis at each step), actions taken, and key performance indicators, including but not limited to:
\begin{itemize}
    \item Total score
    \item Success or failure
    \item Remaining lives
    \item Number of coins collected
    \item Exploration rate
    \item Number of monsters defeated (Level 3 only)
    \item Number of obstacles destroyed (Level 3 only)
    \item Total steps taken
\end{itemize}
For example, after the agent’s initial attempt on the third map of Level 2, the system records both the reasoning process and performance metrics. The reflection module then summarizes the strengths and weaknesses of this attempt, forming the ``subjective memory'' for that level.

\begin{figure}[h]
\centering
\hspace*{-0.5cm} % 向左移动 2 厘米
\includegraphics[scale=0.45]{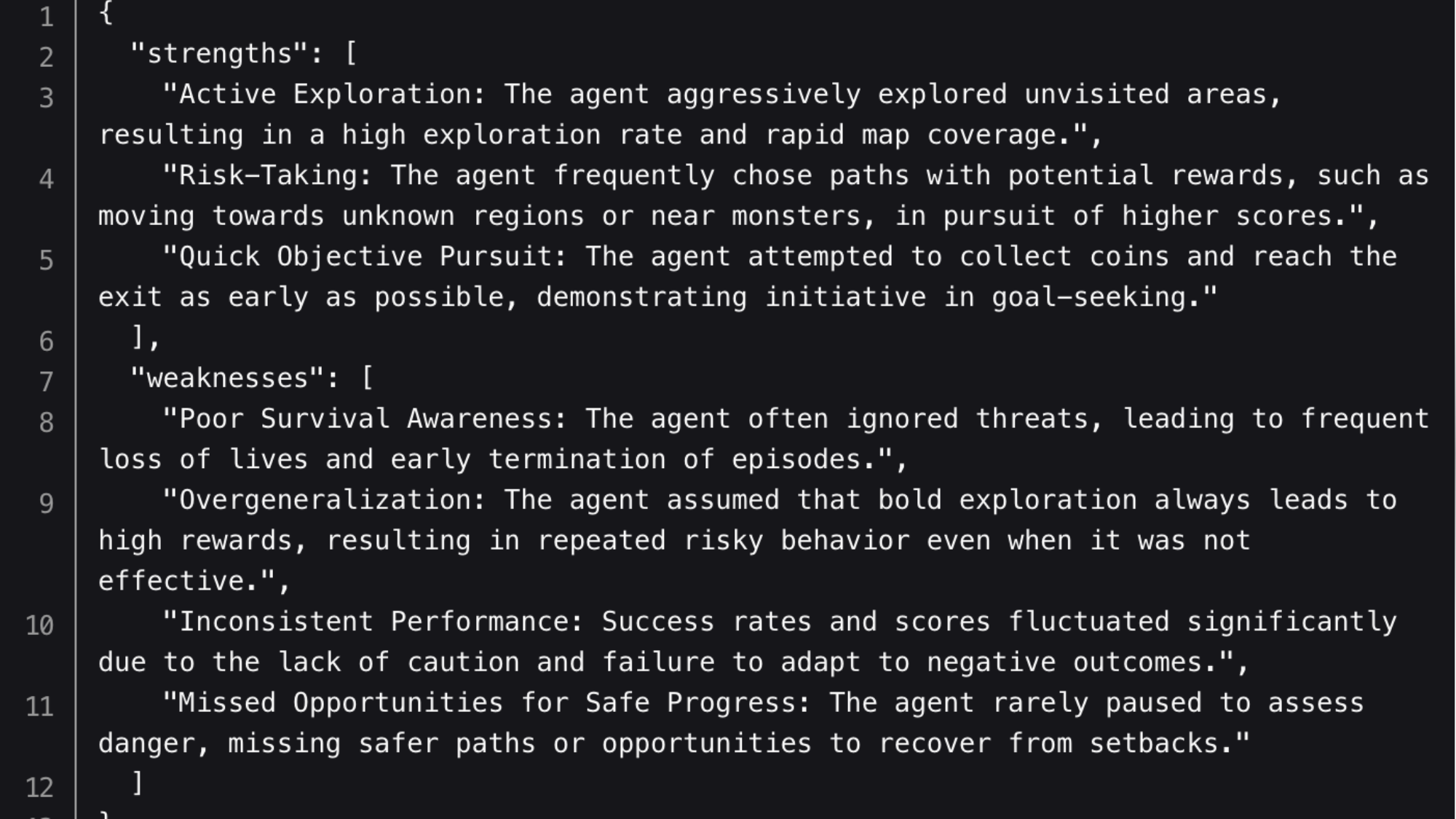}
\caption{~\textbf{Subjective experience guided by adventure} }
\label{fig:purse_profit_experience.pdf}
\end{figure}

\begin{figure}[h]
\centering
\hspace*{-0.5cm} % 向左移动 2 厘米
\includegraphics[scale=0.45]{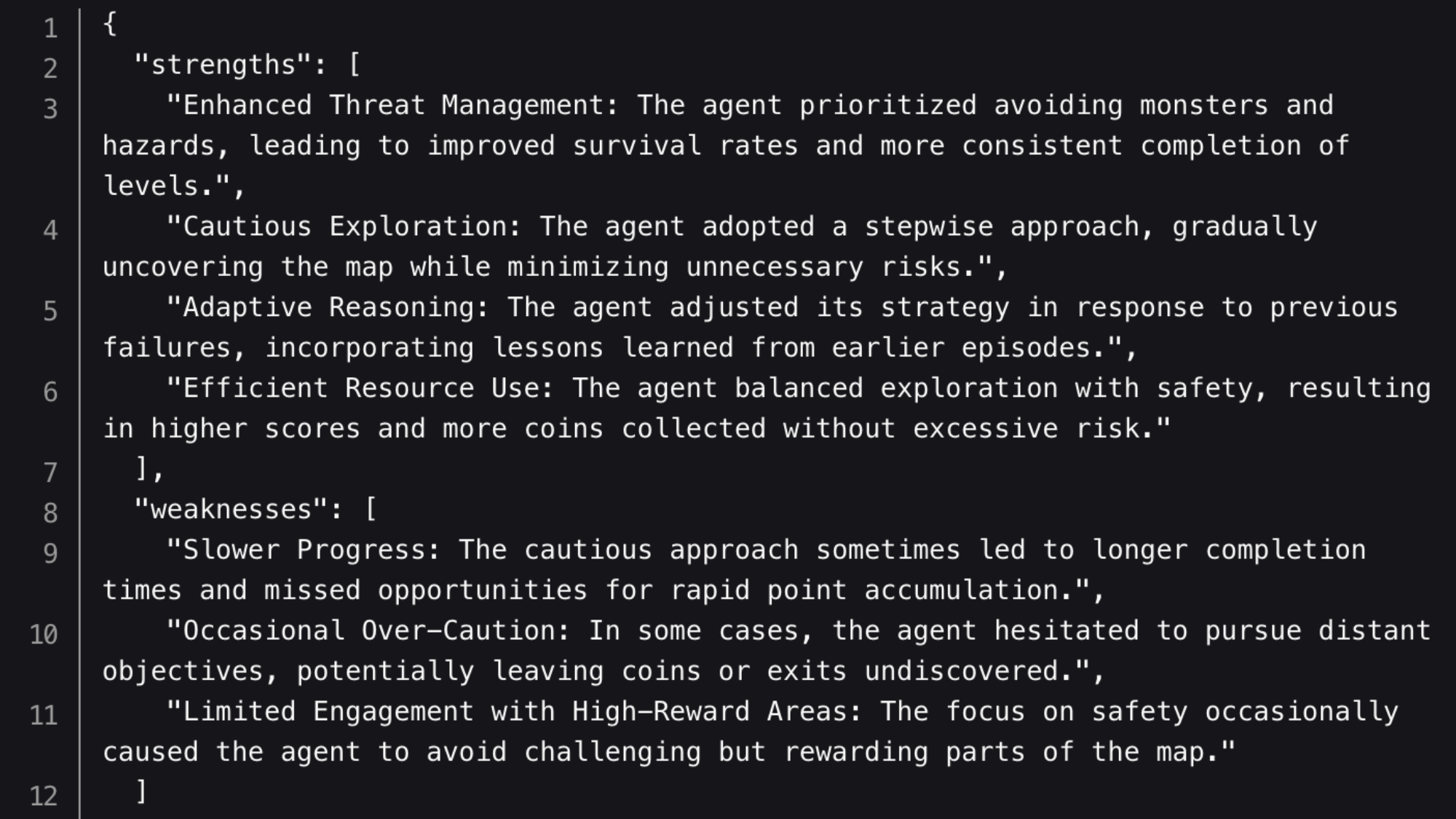}
\caption{~\textbf{Subjective experience of survival orientation} }
\label{fig:safe_experience.pdf}
\end{figure}

\textbf{Verification of Subjective Experience}

The agent then reattempts the same level, this time guided by the previously summarized strengths and weaknesses. The system again records the performance metrics. If there is a significant improvement (e.g., higher score, successful completion), the subjective memory is considered effective and promoted to ``truth knowledge,'' which is added to a global knowledge base for future reference.

\textbf{Induction and Consolidation of Truths}

As training progresses, the truth knowledge base accumulates experiences from different levels and episodes. To prevent redundancy, the system periodically consolidates the knowledge base by:
\begin{itemize}
    \item Removing duplicate entries
    \item Merging similar experiences with different expressions into more representative, generalized knowledge points
    \item Retaining key information from original entries to preserve detail
\end{itemize}
For instance, two similar entries regarding monster avoidance may be merged into a single, more comprehensive guideline.

\vspace{0.5em} \noindent The evolution of generation-oriented truths in the agent’s learning process is clearly reflected in Figure~\ref{fig:purse_profit_experience.pdf} and Figure~\ref{fig:safe_experience.pdf}.

In the early stage (see Figure~\ref{fig:purse_profit_experience.pdf}), the agent’s subjective experience was characterized by a strong tendency toward risk and reward. For instance, the agent frequently prioritized “Active Exploration” and “Risk-Taking,” aggressively venturing into unvisited or potentially dangerous areas in pursuit of higher scores. This is further evidenced by the “Quick Objective Pursuit,” where the agent attempted to collect coins and reach the exit as soon as possible. However, such strategies often led to “Poor Survival Awareness,” with the agent ignoring threats and losing all lives early in the episode. The tendency to “Overgeneralize” from occasional successes—assuming that bold exploration always yields high rewards—resulted in inconsistent performance and frequent failures, as the agent did not sufficiently adapt its strategy after negative outcomes.

After several rounds of iterative learning and reflection, the agent’s approach shifted significantly, as illustrated in Figure~\ref{fig:safe_experience.pdf}. At this stage, the agent’s subjective experience emphasized “Enhanced Threat Management” and “Cautious Exploration.” The agent learned to prioritize survival by actively avoiding monsters and hazards, and adopted a more stepwise, risk-averse approach to map exploration. Notably, the agent demonstrated “Adaptive Reasoning,” adjusting its strategy based on previous failures—such as learning to avoid repeating risky moves that previously led to loss of lives. This resulted in “Efficient Resource Use,” with the agent achieving higher scores and collecting more coins while maintaining safety. Although this cautious approach sometimes led to “Slower Progress” and occasional missed opportunities for rapid point accumulation, it marked a clear improvement in the agent’s ability to balance exploration with survival, leading to more consistent and successful task completion.

In summary, the transition from the early, risk-seeking strategy to a more survival-oriented approach demonstrates the effectiveness of the ExpVer-Agent mechanism in enabling the agent to refine its decision-making process, ultimately achieving a better balance between exploration and safety in complex environments.

\section*{Appendix B: Prompts Used in EEB Games}

This appendix provides the detailed prompt designs used in the two GAME modules of the EEB system: (1) prompts for agent-environment interaction, and (2) prompts for the ExpVer (Experience Verification) module. These prompts are critical for guiding the large language model (LLM) to perform effective reasoning, decision-making, and self-reflection in the maze navigation tasks.

\subsection*{B.1 Agent-Environment Interaction Prompts}

For both GAME 1 and GAME 2, the agent interacts with the maze environment through carefully constructed prompts. The prompts are designed to provide the LLM with comprehensive information about the current game state, available actions, and key objectives, ensuring that the agent can make informed and context-aware decisions.

\textbf{Example Prompt for Maze Navigation Game:}
\begin{verbatim}
You are an intelligent agent solving a maze problem. 
Your task is to navigate through the maze efficiently while collecting rewards and avoiding
dangers.

Core Game Elements:
- A: Your current position
- G: Goal (always visible)
- C: Coin (+500 points)
- #: Wall (costs life if hit)
- ?: Unexplored area
- .: Empty space

Your Priorities (in order):
1. Stay alive (avoid walls/monsters)
2. Reach the goal
3. Collect coins when safe
4. Explore efficiently
5. Minimize steps

You will receive the current game state and must choose an action (0-11) based on careful
analysis of the situation.
Always explain your reasoning before making a decision.

Level 1 Characteristics:
- 9x9 grid size
- No monsters
- 5 coins to collect
- Focus on basic navigation and coin collection

Level 2 Characteristics:
- 9x9 grid size
- Contains monsters (M) that move randomly
- 5 coins to collect
- Requires careful planning to avoid monsters

Level 3 Characteristics:
- 9x9 grid size
- Contains monsters and special items
- Items available:
  * T: Shovel (break walls, 3 uses)
  * W: Sword (defeat monsters)
  * N: Magnet (attract nearby coins)
  * K: Key (required for goal)
- Most complex navigation

Training Knowledge (use these insights to make better decisions):
1. [knowledge 1]
2. [knowledge 2]
... (Each item from truth_knowledge, one per line)

Current Game State:
[Map string generated by MapProcessor.format_map_for_agent(), e.g.]:
   0 1 2 3 4 5 6 7 8
  ---------------------
 0 | . . . ? ? ? ? ? ?
 1 | . A . ? ? ? ? ? ?
 2 | . . . ? ? ? ? ? ?
 3 | . . . ? ? ? ? ? ?
 4 | . . . ? ? ? ? ? ?
 5 | . . . ? ? ? ? ? ?
 6 | . . . ? ? ? ? ? ?
 7 | . . . ? ? ? ? ? ?
 8 | . . . ? ? ? ? ? ?

Current position (row,col): (1,1)

[If full vision:]
NOTE: You have full vision of the entire map. You can see all obstacles, coins, monsters, 
and items without the need to explore.

Game Status:
- Score: [current score]
- Lives: [current lives]
- Current Position (row,col): ([row],[col])

Movement System:
- Actions 0-2: Move UP (row-1) [1/2/3 steps]
- Actions 3-5: Move DOWN (row+1) [1/2/3 steps]
- Actions 6-8: Move LEFT (col-1) [1/2/3 steps]
- Actions 9-11: Move RIGHT (col+1) [1/2/3 steps]

Available Actions:
- 0: Move UP 1 step
- 1: Move UP 2 steps
- 2: Move UP 3 steps
- 3: Move DOWN 1 step
- 4: Move DOWN 2 steps
- 5: Move DOWN 3 steps
- 6: Move LEFT 1 step
- 7: Move LEFT 2 steps
- 8: Move LEFT 3 steps
- 9: Move RIGHT 1 step
- 10: Move RIGHT 2 steps
- 11: Move RIGHT 3 steps

[If Level 3, add item status:]
Current items status:
- Shovel: Equipped (Uses remaining: [number]) / Not equipped
- Sword: Equipped / Not equipped
- Magnet: Equipped / Not equipped
- Key: Collected / Not collected (required to finish)

Scoring System:
- New cell explored: +10 points
- Coin collected: +500 points
- Step taken: -50 points
- Life lost: -1000 points
- Goal reached: +2000 points

Please analyze the current situation and choose your next action:
1. Analyze visible area and potential risks
2. Consider exploration value and rewards
3. Choose action number (0-11)

[If Level 2 or 3:]
- You cannot touch monsters (M), or you'll lose a life and return to start

[If Level 3:]
Special rules for Level 3:
- You must collect the key (K) before you can enter the goal
- With a shovel (T), you can break through walls without losing lives (3 uses)
- With a sword (W), you can defeat monsters without losing lives
- With a magnet (N), you can collect coins in a 5x5 area around you

The ultimate goal is to explore the map, collect coins, reach the goal, while maintaining a
high score

Response format:
1. First analyze the current situation, including explored areas, coin positions, and
potential risks
2. Consider possible movement options and their consequences, especially focusing on 
exploration value
3. Finally, provide your choice using the format "Action: X" where X is a number 
between 0-
11

For example:
After analysis, you should write "Action: 9" to indicate moving right by 1 step
\end{verbatim}

\textbf{Example Prompt for Match-2 Game:}
\begin{verbatim}
system_prompt = """
You are an AI assistant for an 8x8 match game (gridSize = 8). The board is an
8x8 grid with colors A, B, C, D, or null (empty). Rules:
- Eliminate \geq2 connected same-color tiles (horizontal/vertical), score = tiles * 
5 + 3 * max(0, tiles - 2).
- Props (each usable once): row (clear row, 32 points), col (clear column, 32 
points), bomb (clear 3x3 area, 12 points), hammer (clear 1 tile, 4 points).
- Each action costs 1 step. Goal: Clear the level by meeting color elimination 
targets (A, B, C, D) within limited steps while maximizing score and minimizing 
steps used.
- Primary objective: Ensure level completion by achieving all color targets.
- Secondary objectives: Maximize total score by prioritizing larger tile 
eliminations and efficient prop usage; minimize steps to preserve remaining 
steps.
- After elimination, new random tiles (A, B, C, D) fall from the top to fill
empty spaces.
- Input: Board (8x8, A/B/C/D/null), score, steps remaining, inventory 
(row/col/bomb/hammer), color targets, current color counts.
- Output: Best action in JSON: {"action": {"type": 
"eliminate"|"row"|"col"|"bomb"|"hammer", "pos": [i,j] (for 
eliminate/bomb/hammer, 0\leqi,j<8), "index": k (for row/col, 0\leqk<8)}}.
- If no valid action, return {"action": null}.
"""
\end{verbatim}

\subsection*{B.2 ExpVer Module Prompts}

The ExpVer (Experience Verification) module is responsible for guiding the agent in self-reflection, hypothesis testing, and truth induction. The prompts for this module are designed to elicit a structured analysis of the agent's previous actions, outcomes, and reasoning, enabling the system to summarize strengths, weaknesses, and actionable insights.

\textbf{Example ExpVer Prompt:}
\begin{verbatim}
# Game Session Analysis

Please analyze the following game session and provide insights about the agent's
performance.

## Game Metrics:
- [key1]: [value1]
- [key2]: [value2]
...

## Session Highlights:

Step 1 (Beginning):
Agent's reasoning: [first completion text]...

Step 2 (Beginning):
Agent's reasoning: [second completion text]...

...

Step N (End):
Agent's reasoning: [last completion text]...

## Analysis Tasks:

1. List the strengths demonstrated in this session. Provide as many as you can 
identify.

2. List the weaknesses or areas for improvement from this session. Provide as 
many as you can identify.

Please format your response as follows:

Strengths:
- [Strength 1]
- [Strength 2]
- [Strength 3]
...

Weaknesses:
- [Weakness 1]
- [Weakness 2]
- [Weakness 3]
...
\end{verbatim}

\begin{verbatim}
# Game Truth Knowledge Organization Task

Please review and organize the following game truth knowledge entries. This is a
progressive knowledge organization process to identify and remove duplicates 
while considering merging highly similar entries.

## Current Knowledge Entries:
1. [Knowledge 1] (Source: ...)
2. [Knowledge 2] (Source: ...)
...

## Organization Requirements:
1. Identify and remove completely duplicate knowledge entries
2. For knowledge entries that are highly similar in meaning but different in
expression, merge them into a more comprehensive entry
3. When merging, preserve the specificity and core meaning of the original 
knowledge, don't lose key details
4. Merged entries should be concise but not at the expense of important 
information
5. If two knowledge points only have minimal similarities, keep them as separate 
entries
6. Knowledge entries without clear similarities should remain unchanged

## Please return the organized knowledge base in the following format:
[Organized knowledge entry 1]
[Organized knowledge entry 2] ...

Note: This is a progressive knowledge organization process, you do not need to 
force a reduction in the number of entries. Only merge or remove entries when 
there is genuine high similarity or duplication.
\end{verbatim}

These prompt templates ensure that the LLM agent can both interact effectively with the environment and perform meaningful self-assessment, supporting the iterative learning and truth induction process central to the EEB framework.

\section*{Appendix C. Example Agent Reasoning Trace (Maze Game, Level 1, GPT-4.1)}

This appendix demonstrates the actual reasoning process of the LLM agent (GPT-4.1) in the first two steps of Level 1 in GAME1. Each step includes the system prompt, the game state, and the model's full reasoning and decision.

\subsection*{Step 0}

\textbf{System Prompt:}
\begin{verbatim}
You are an intelligent agent solving a maze problem. Your task is to navigate through the
maze efficiently while collecting rewards and avoiding dangers.

Core Game Elements:
- A: Your current position
- G: Goal (always visible)
- C: Coin (+500 points)
- #: Wall (costs life if hit)
- ?: Unexplored area
- .: Empty space

Your Priorities (in order):
1. Stay alive (avoid walls/monsters)
2. Reach the goal
3. Collect coins when safe
4. Explore efficiently
5. Minimize steps

You will receive the current game state and must choose an action (0-11) based on careful
analysis of the situation.
Always explain your reasoning before making a decision.
Level 1 Characteristics:
- 7x7 grid size
- No monsters
- 5 coins to collect
- Focus on basic navigation and coin collection
\end{verbatim}

\textbf{Game State (partial):}
\begin{verbatim}
Current Game State:
Game Coordinate System (row,col):
- Rows: 0 (top) to N (bottom)
- Cols: 0 (left) to N (right)

   0  1  2  3  4  5  6  7  8
  ----------------------------
 0 | ? ? ? ? ? ? ? ? ?
 1 | ? ? ? ? ? ? ? ? ?
 2 | ? ? ? ? ? ? ? ? ?
 3 | ? ? ? ? ? ? ? ? ?
 4 | ? ? ? ? ? ? ? ? ?
 5 | ? ? ? ? ? ? ? ? ?
 6 | ? ? ? ? ? ? ? ? ?
 7 | . . ? ? ? ? ? ? ?
 8 | A . ? ? ? ? ? ? ?

Current position (row,col): (8,0)
\end{verbatim}

\textbf{Model Reasoning and Decision:}
\begin{verbatim}
### 1. Situation Analysis

- **Visible Area:**  
  The only explored cells are:
  - (8,0): Agent's current position (A)
  - (7,0): Explored (.)
  - (7,1): Explored (.)
  - (8,1): Explored (.)

  All other cells are unexplored (?).

- **Coins:**  
  No coins are currently visible.

- **Obstacles/Monsters:**  
  None are visible.

- **Risks:**  
  No immediate risks. All adjacent cells that are explored are safe (empty).

### 2. Movement Options & Consequences

- **Possible Moves:**  
  From (8,0), the agent can move:
  - UP to (7,0): Already explored, no exploration points, but safe.
  - RIGHT to (8,1): Already explored, no exploration points, but safe.
  - Any other direction (LEFT or DOWN) would go out of bounds (since (8,0) is at the
  bottom-left corner), so those are not valid.

- **Exploration Value:**  
  To maximize exploration (and thus score), the agent should move into unexplored 
  territory.  
  The only way to do this is to move UP or RIGHT first (to a safe, explored cell), then 
  from there, move into unexplored areas.

- **Step Minimization:**  
  Since both (7,0) and (8,1) are equally valid, but both are already explored, the agent 
  will not gain exploration points for the next move. However, moving to (8,1) (RIGHT) may
  open up more unexplored territory to the right, while moving UP to (7,0) may open up 
  more unexplored territory above.

- **Preference:**  
  Both options are equally safe and equally efficient at this stage. Either is acceptable,
  but moving RIGHT (Action 9) may be slightly preferable, as it keeps the agent on the 
  bottom row, potentially allowing for more lateral exploration before moving up.

### 3. Decision

**Action: 9** (Move RIGHT 1 step to (8,1))

**Reasoning:**  
- Safe move (no risk)
- Prepares for further exploration into the right side of the map
- Keeps options open for exploring both up and right in future moves

---

**Action: 9**
\end{verbatim}

\subsection*{Step 1}

\textbf{System Prompt:}
\begin{verbatim}
You are an intelligent agent solving a maze problem. Your task is to navigate through the 
maze efficiently while collecting rewards and avoiding dangers.

Core Game Elements:
- A: Your current position
- G: Goal (always visible)
- C: Coin (+500 points)
- #: Wall (costs life if hit)
- ?: Unexplored area
- .: Empty space

Your Priorities (in order):
1. Stay alive (avoid walls/monsters)
2. Reach the goal
3. Collect coins when safe
4. Explore efficiently
5. Minimize steps

You will receive the current game state and must choose an action (0-11) based on careful 
analysis of the situation.
Always explain your reasoning before making a decision.
Level 1 Characteristics:
- 7x7 grid size
- No monsters
- 5 coins to collect
- Focus on basic navigation and coin collection
\end{verbatim}

\textbf{Game State (partial):}
\begin{verbatim}
Current Game State:
Game Coordinate System (row,col):
- Rows: 0 (top) to N (bottom)
- Cols: 0 (left) to N (right)

   0  1  2  3  4  5  6  7  8
  ----------------------------
 0 | ? ? ? ? ? ? ? ? ?
 1 | ? ? ? ? ? ? ? ? ?
 2 | ? ? ? ? ? ? ? ? ?
 3 | ? ? ? ? ? ? ? ? ?
 4 | ? ? ? ? ? ? ? ? ?
 5 | ? ? ? ? ? ? ? ? ?
 6 | ? ? ? ? ? ? ? ? ?
 7 | . . # ? ? ? ? ? ?
 8 | ? A C ? ? ? ? ? ?

Current position (row,col): (8,1)
\end{verbatim}

\textbf{Model Reasoning and Decision:}
\begin{verbatim}
### 1. Situation Analysis

- **Current Position:** (8,1)
- **Visible Map:** Only the bottom two rows (rows 7 and 8) are partially visible.
    - (8,2): Coin (C)
    - (7,2): Wall (#)
    - (7,0), (7,1): Empty (.)
    - All other cells: Unexplored (?)
- **Obstacles:** Wall at (7,2)
- **Coins:** Only one visible coin at (8,2), directly to the right of the agent.
- **No monsters** are present.
- **Goal (G):** Not yet visible.

**Risks:**
- The only immediate risk is the wall at (7,2). Moving up from (8,2) would hit the wall,
but moving right from (8,1) to (8,2) is safe.

### 2. Movement Options and Consequences

#### Immediate Options:
- **Move RIGHT (Action 9):** Move to (8,2), collect the coin (+500), explore a new cell 
(+10), cost 1 step (-50). Net gain: +460 points. No risk.
- **Move UP (Action 0):** Move to (7,1), explore a new cell (+10), cost 1 step (-50). No 
immediate reward, but opens up exploration vertically.
- **Move LEFT (Action 6):** Move to (8,0), unexplored cell, similar exploration value as 
moving up.
- **Move UP 2/3, LEFT 2/3, RIGHT 2/3:** All would move into unexplored territory, but 
since we know there is a coin directly to the right, the most efficient and rewarding move 
is to collect it first.

#### Strategic Considerations:
- **Priority 1:** Stay alive (no risk in moving right).
- **Priority 2:** Reach the goal (not visible yet, so explore efficiently).
- **Priority 3:** Collect coins (coin is immediately available).
- **Priority 4:** Explore efficiently (moving right will also reveal more of row 8 and 
possibly row 7).
- **Priority 5:** Minimize steps (moving right is a single step and optimal for reward).

### 3. Decision

The best move is to collect the coin at (8,2) by moving right 1 step. This is safe, 
maximizes immediate reward, and continues exploration.

**Action: 9**
\end{verbatim}

\section*{Appendix D. Example Agent Reasoning Trace (Match-2 Game, esay, GPT-4.1)}

This appendix demonstrates the actual reasoning process of the LLM agent (GPT-4.1) in the 
first several steps of Level 1 in Match-2 Game (easy). Each step includes the board state,
action, and the model's full reasoning and decision.

\subsection*{Step 1}

\textbf{Board State (partial):}
\begin{verbatim}
C D D A B B C B
C C C B A A B C
A B B C B D B D
C C B A B B A B
B D C B A D A B
A B B D A C A A
D B A C C D B D
D B A A A C B D
Score: 0, Steps remaining: 15
Inventory: row=0, col=1, bomb=2, hammer=1
Color targets: A:10, B:6, C:8, D:6
Current counts: A:10, B:7, C:11, D:6
\end{verbatim}

\textbf{Action:} \texttt{eliminate at (4,3)}

\textbf{Model Reasoning:}
\begin{verbatim}
Let's analyze the board and objectives:

**Color targets:** {"A": 10, "B": 6, "C": 8, "D": 6}  
**Current counts:** All zero  
**Steps left:** 15  
**Props:** row (1), col (2), bomb (2), hammer (1)  
**Goal:** Maximize score, minimize steps, guarantee level completion.

### Step 1: Find the largest connected group (\geq2) of any color

#### Scan for large groups (to maximize score):

- **Row 0:**  
  - "A" at (0,3)-(0,4)-(0,6)-(0,7): 4 connected "A"s (positions [0,3],[0,4],[0,6],[0,7]),
  but not all adjacent.  
  - "A" at (0,3)-(0,4): 2 connected  
  - "A" at (0,6)-(0,7): 2 connected  
- **Row 1:**  
  - "B" at (1,2)-(1,3): 2 connected  
  - "B" at (1,5)-(1,6): 2 connected  
- **Row 3:**  
  - "B" at (3,0)-(3,1): 2 connected  
- **Row 4:**  
  - "B" at (4,3)-(4,4)-(4,5): 3 connected  
- **Row 5:**  
  - "C" at (5,0)-(5,1): 2 connected  
- **Row 6:**  
  - "D" at (6,3)-(7,3): vertical, but not connected  
  - "B" at (6,1)-(6,2): 2 connected  
- **Row 7:**  
  - "B" at (7,1)-(7,6): not connected  
  - "D" at (7,0)-(7,2): not connected  

#### Largest group:  
- "B" at (4,3)-(4,4)-(4,5): 3 connected  
- "A" at (0,3)-(0,4): 2 connected  
- "A" at (0,6)-(0,7): 2 connected  
- "B" at (1,2)-(1,3): 2 connected  
- "B" at (1,5)-(1,6): 2 connected  
- "B" at (3,0)-(3,1): 2 connected  
- "C" at (5,0)-(5,1): 2 connected  
- "B" at (6,1)-(6,2): 2 connected  

### Step 2: Evaluate prop usage

- **Row/Col:** No row or column has a huge concentration of a single color or a large 
number of target colors.
- **Bomb:** No 3x3 area contains a large group of a single color.
- **Hammer:** Not efficient for first move.

### Step 3: Best elimination for targets and score

- **Eliminating (4,3) "B" group (3 tiles):**
  - Score: 3*5 + 3*(3-2) = 15 + 3 = 18
  - Progress: 3/6 "B" target
  - Efficient: Yes, largest group, helps with "B" target

### Step 4: Output

**Best action:** Eliminate the 3-"B" group at (4,3)

**JSON:**
{
  "action": {
    "type": "eliminate",
    "pos": [4, 3]
  }
}

**Rationale:**  
- Largest available group (3 tiles) for maximum score and target progress.
- Efficient use of step.
- No prop needed yet; save for later if targets become hard to reach.

\end{verbatim}

% --------------------------------------------------------

\subsection*{Step 2}

\textbf{Board State (partial):}
\begin{verbatim}
C D D A B B C B
C C C B A A B C
A B B C B D B D
C C B A B B A B
B D C B A D A B
A B B D A C A A
D B A C C D B D
D B A A A C B D
Score: 34, Steps remaining: 14
Inventory: row=0, col=1, bomb=2, hammer=1
Color targets: A:10, B:6, C:8, D:6
Current counts: A:10, B:7, C:11, D:6
\end{verbatim}

\textbf{Action:} \texttt{eliminate at (0,3)}

\textbf{Model Reasoning:}
\begin{verbatim}
Given your board, targets, and inventory, let's analyze the best action:

### 1. Primary Objective: Level Completion
- You need: A: 10, B: 1, C: 8, D: 6 (since current counts are A:0, B:5, C:0, D:0 and 
targets are A:10, B:6, C:8, D:6).
- Steps left: 14 (ample, but efficiency is still important).

### 2. Secondary Objectives
- Maximize score: Prefer larger connected groups.
- Minimize steps: Prefer actions that clear more target tiles per step.
- Prop usage: Use props if they clear many targets or help with hard-to-reach tiles.

---

### 3. Board Analysis: Find Largest Groups of Target Colors

#### Scan for large groups (\geq2) of A, B, C, D:

#### "A" Groups:
- [0,3]-[0,4]-[0,5]-[0,6]-[0,7] (Row 0, columns 3-7): 5 connected A's (0,3)-(0,7) 
(horizontal).
- [1,4]-[1,5] (Row 1, columns 4-5): 2 connected A's.
- [2,2]-[2,3] (Row 2, columns 2-3): 2 connected A's.
- [4,4]-[4,6] (Row 4, columns 4 and 6): not connected.
- [5,4]-[5,6]-[5,7] (Row 5, columns 4,6,7): 2 connected (5,6)-(5,7).

#### "B" Groups:
- [3,0]-[3,1] (Row 3, columns 0-1): 2 connected B's.
- [3,4]-[3,5]-[3,7] (Row 3, columns 4,5,7): 2 connected (4-5), (5-7) not connected.
- [6,1]-[6,2] (Row 6, columns 1-2): 2 connected B's.
- [7,1]-[7,6] (Row 7, columns 1 and 6): not connected.

#### "C" Groups:
- [0,0]-[1,1]-[2,0] (not connected).
- [2,4]-[2,5]-[2,6] (Row 2, columns 4-6): 3 connected C's.
- [5,1]-[5,3]-[5,5] (Row 5, columns 1,3,5): not connected.
- [7,5] (single).

#### "D" Groups:
- [2,1]-[2,7] (not connected).
- [4,2]-[4,3] (Row 4, columns 2-3): 2 connected D's.
- [6,0]-[7,0] (Column 0, rows 6-7): 2 connected D's.
- [6,3]-[6,5] (Row 6, columns 3,5): not connected.

---

### 4. Best Action Candidates

#### Largest group:
- "A": 5 connected at [0,3]-[0,7] (horizontal).
- "C": 3 connected at [2,4]-[2,6] (horizontal).
- "B": Only 2 connected at best.
- "D": Only 2 connected at best.

#### Score for 5 A's: 5*5 + 3*(5-2) = 25 + 9 = 34 points.

#### This move also makes a big dent in the A target (5/10 in one move).

---

### 5. Prop Consideration
- Row/Col clear: Would clear 8 tiles, but not all would be targets, and you have only 1 
row clear.
- Bomb: 3x3 area, but no area covers more than 5 A's or more targets than the above group.
- Hammer: Only for single tiles, not efficient here.

---

### 6. Recommendation

**Best move:** Eliminate the 5 connected "A" tiles at the top row, starting at [0,3].

**Action:**
{"action": {"type": "eliminate", "pos": [0,3]}}

- This clears 5/10 A's in one step, scores 34 points, and may set up further matches as 
new tiles fall.
- It is the most efficient move for both level completion and score maximization at this
step.

\end{verbatim}